\documentclass[11pt]{article}

\usepackage[final]{acl}

\usepackage{times}
\usepackage{latexsym}

\usepackage[T1]{fontenc}
\usepackage[utf8]{inputenc}

\usepackage{microtype}

\usepackage{inconsolata}

\usepackage{graphicx}
\usepackage{amsmath}
\usepackage{amssymb}
\usepackage{float}
\usepackage{tcolorbox}
\tcbuselibrary{breakable}
\usepackage{booktabs}
\usepackage{multirow}
\usepackage{makecell}
\usepackage[table]{xcolor}
\definecolor{steerbg}{RGB}{230,242,255}
\definecolor{headbg}{RGB}{245,245,245}   
\definecolor{goodgreen}{HTML}{2CA02C}
\definecolor{badred}{HTML}{D62728}
\usepackage{subcaption}
\usepackage{graphicx}
\newcommand{\ourmethod}{LSS-CoT} % or \textsc{Lss-CoT}

\title{Discovering a Shared Logical Subspace: Steering LLM Logical Reasoning via Alignment of Natural-Language and Symbolic Views}

\author{\vspace{8pt}
Feihao Fang\textsuperscript{1} \hspace{2em} My T. Thai\textsuperscript{2} \hspace{2em} Yuanyuan Lei\textsuperscript{2}\\
       \textsuperscript{1}University of Illinois Urbana-Champaign, Champaign, IL\\ \textsuperscript{2}Computer \& Information Science and Engineering, University of Florida, Gainesville, FL\\
       \texttt{feihaof2@illinois.edu}, \texttt{yuanyuan.lei@ufl.edu}}

\begin{document}
\maketitle
\begin{abstract}
%Large language models (LLMs) often falter on multi-step logical reasoning. Existing approaches typically either operate purely in the natural-language (NL) view or attach symbolic structure as an external module. We instead ask whether LLMs contain a shared internal ``logical subspace'' that simultaneously aligns NL and symbolic proofs. Using paired residual activations from NL and symbolic proofs, we apply Canonical Correlation Analysis (CCA) to learn a low-dimensional subspace with maximum cross-view correlation, and introduce a training-free method that steers Chain-of-Thought (CoT) decoding along this subspace. On four logical reasoning benchmarks and five open-weight LLMs, steering along this subspace improves CoT accuracy by up to 11 absolute points and transfers from synthetic proofs to out-of-domain logical reasoning benchmarks. Finally, analysis of activations shows that this subspace organizes interpretable logical roles and that projection energy yields an unsupervised signal of reasoning correctness.

Large Language Models (LLMs) still struggle with multi-step logical reasoning. Existing approaches either purely refine the reasoning chain in natural language form or attach a symbolic solver as an external module. In this work, we instead ask whether LLMs contain a shared internal \textit{logical subspace} that simultaneously aligns natural-language and symbolic-language views of the reasoning process. Our hypothesis is that this logical subspace captures logical reasoning capabilities in LLMs that are shared across views while remaining independent of surface forms. To verify this, we employ Canonical Correlation Analysis on the paired residual activations from natural-language and symbolic-language reasoning chains, learning a low-dimensional subspace with maximum cross-view correlation. Furthermore, we design a training-free approach that steers LLMs reasoning chain along this logical subspace, thereby leveraging the complementary reasoning signals from both views. Experiments on four logical reasoning benchmarks demonstrate the effectiveness of our approach, improving accuracy by up to 11 percentage points and generalizing well on out-of-domain problems\footnote{The code and data link is: \url{https://github.com/lei-nlp-lab/logical_subspace_acl_2026}}.

\end{abstract}

\section{Introduction}

%Large language models (LLMs) still struggle with multi-step logical reasoning \cite{xu2025logicalreasoning}. Even with Chain-of-Thought (CoT) prompting, they often produce fluent yet logically invalid “proofs”, especially on multi-hop rule-based reasoning benchmarks \cite{turpin2023languagemodelsdontsay}.

Logical reasoning in LLMs refers to their ability to follow rules, connect premises to conclusions, and carry out multi-step inferences \cite{liu2023evaluating}. Despite recent advances in natural-language understanding, LLMs still fall short on complex, multi-step logical reasoning problems \cite{xu2025logicalreasoning}. Strong logical reasoning capability is important for applications that require multi-step decision making, such as math, scientific analysis, planning, coding etc. \cite{cheng2025empowering}. Thus, improving LLM logical reasoning is a critical problem.

%Prior work improves LLM reasoning along three main directions. Prompting- and training-based approaches refine prompts and decoding or train on natural-language CoT traces to elicit better step-by-step reasoning \cite{wei2022cot,wang2023selfconsistencyimproveschainthought,zhou2023leasttomostpromptingenablescomplex,zou2025reasonfluxprmtrajectoryawareprmslong}. Neuro-symbolic methods inject structure by integrating logical constraints into training objectives or coupling LLMs with external theorem provers and verifiers \cite{xu-etal-2024-faithful,pan-etal-2023-logic}. Activation-steering methods learn directions in internal activation spaces and intervene linearly at inference time \cite{chen2025sealsteerablereasoningcalibration,azizi2025activationsteeringchainofthoughtcompression,tang-etal-2025-unlocking}. However, these approaches either operate purely in the natural-language view or treat symbolic components as external modules; none explicitly learns a shared internal representation jointly aligned between natural-language and symbolic views.

Prior works can be broadly grouped into two streams: natural-language-dependent methods and neural-symbolic methods. The first line of work uses prompting or training-based techniques to optimize natural-language chain-of-thought traces during decoding or training \cite{wei2022cot, zou2025reasonfluxprmtrajectoryawareprmslong}. Another line of neural-symbolic approaches augment LLMs with external symbolic provers or verifiers \cite{pan-etal-2023-logic,lei-huang-2024-boosting}. However, these methods either focus solely on refining reasoning in natural-language form or rely on externally attached symbolic components. In this work, we introduce a new framework that aligns natural-language and symbolic views of the logical reasoning process, thus leveraging complementary reasoning signals from both views.

\begin{figure}[t]
\small
\centering
\begin{tabular}{@{}p{0.48\linewidth}p{0.48\linewidth}@{}}
\toprule
\textbf{Natural-Language Proof} & \textbf{Symbolic-Language Proof} \\
\midrule
\begin{minipage}[t]{\linewidth}
\textbf{Context.} Every bird can fly. Tweety is a bird.\\
\textbf{Claim.}\\ Tweety can fly.\\
\textbf{Proof.}\\
Tweety is a bird.\\
All birds can fly.\\
Therefore, Tweety can fly.
\end{minipage}
&
\begin{minipage}[t]{\linewidth}
\textbf{Facts.}\\
\(\mathrm{bird}(\mathrm{tweety})\)\\
\mbox{$\forall x\, (\mathrm{bird}(x) \rightarrow \mathrm{can\_fly}(x))$}\\[0.4em]
\textbf{Goal.}\\
\mbox{$\mathrm{can\_fly}(\mathrm{tweety})$}\\[0.4em]
\textbf{Proof.}\\
\(\mathrm{bird}(\mathrm{tweety})\).\\
\mbox{$\forall x\, (\mathrm{bird}(x) \rightarrow \mathrm{can\_fly}(x))$}.\\
\(\mathrm{can\_fly}(\mathrm{tweety})\).
\end{minipage}
\\
\bottomrule
\end{tabular}
\caption{An example of a logical reasoning chain expressed in natural language and symbolic language.}

\label{fig:two-views}
\end{figure}

%Many logical reasoning problems naturally admit two complementary views of the same underlying derivation: (i) a natural-language proof written as a step-by-step explanation, and (ii) a corresponding symbolic proof expressed as a sequence of rule applications or logical clauses. The two views agree on logical content but differ in surface form: the symbolic proof exposes formal structure, while the NL proof reflects how a model verbalizes that structure. This motivates a core question: \emph{do LLMs form a shared internal “logical” subspace that aligns with both views, and can we exploit it to steer model-generated CoT reasoning at inference time?}

Our idea is inspired by the observation that each logical reasoning problem can be solved in two complementary representations of the same underlying derivation: (i) a natural-language (NL) proof written as a step-by-step verbal explanation, and (ii) a corresponding symbolic proof expressed as a sequence of rules or logical clauses, as illustrated in Figure \ref{fig:two-views}. The two expressions agree on the logical reasoning process but differ in surface form: the symbolic proof exposes formal structure, while the NL proof reflects how a model verbalizes that structure. This motivates our core research question: \emph{do LLMs contain a shared internal logical subspace that aligns with both views, and can we exploit it to enhance LLM logical reasoning?}

%We address this question by hypothesizing that, at selected layers of the residual stream, there exists a low-dimensional \emph{multi-view logical subspace} in which activations from paired natural-language and symbolic proofs of the same instance are strongly aligned. To instantiate this idea, we build a multi-view activation dataset on several logical reasoning benchmarks and apply a multi-view representation learner to extract a low-dimensional subspace whose directions are maximally correlated with the symbolic view. On top of this representation, we introduce a simple, training-free steering method that operates during CoT generation: as the model autoregressively generates each CoT token, we linearly amplify its projection onto the learned subspace, nudging the hidden state toward the shared NL–symbolic representation while leaving model weights unchanged.

We address this question by hypothesizing that there exists a low-dimensional \emph{multi-view logical subspace} in LLMs, in which activations from paired natural-language and symbolic proofs of the same instance are strongly aligned. To instantiate this idea, we employ Canonical Correlation Analysis algorithm \cite{raghu2017svccasingularvectorcanonical} on the paired residual activations from both natural-language and symbolic reasoning chains, learning a shared low-dimensional subspace. By maximizing cross-view correlation, we expect this subspace to capture logical reasoning capabilities in LLMs that are shared across views and independent of surface forms.

%Empirically, steering along this multi-view logical subspace improves reasoning across multiple logical benchmarks and five open-weight LLMs, and the same subspace transfers to out-of-distribution tasks. Compared to neuro-symbolic integrations, our approach requires no extra supervision, no change to the training objective, and no external solver. Compared to prior activation-steering methods, our directions are learned from paired NL and symbolic views and are explicitly optimised for cross-view alignment rather than single-view heuristics.

Furthermore, to fully exploit the complementary reasoning signals contained in the shared logical subspace, we design a training-free approach that nudges the LLM’s reasoning chain at inference time: as the model autoregressively generates each token, we linearly amplify the projection of each token's activation onto the learned subspace, thereby steering the hidden state towards the shared NL-symbolic subspace while leaving model weights unchanged. Our innovation over prior work is that our method moves beyond single-view heuristics and integrates complementary reasoning signals through cross-view alignment, without requiring additional training or external symbolic solvers.

Experiments across four logical reasoning benchmarks and five LLMs demonstrate the effectiveness of our approach, improving accuracy by up to 11\%. The learned logical subspace also generalizes well to out-of-distribution reasoning problems. Moreover, our analysis reveals several insights: (i) the logical subspace in LLMs encodes both semantic and logical structure information (ii) alignment between natural-language and symbolic views strengthens in higher layers of LLMs (iii) projection energy within the learned logical subspace correlates positively with reasoning correctness (iv) steering along the logical subspace increases the use of logical connective words such as \textit{since}, \textit{so}, while decreasing reliance on vague reasoning verbs such as \textit{think}, \textit{know}, \textit{assume}.

Our contributions can be summarized as:
\begin{itemize}
    \item We discover a shared internal \textit{logical subspace} in LLMs that aligns both natural-language and symbolic views of the reasoning process.
    \item We propose a training-free method that steers LLMs’ generation along the logical subspace, thus leveraging cross-view reasoning signals.
    \item We demonstrate up to 11\% improvement on logical reasoning, show strong out-of-domain generalization, and provide analysis revealing the structure and behavior of the subspace.
\end{itemize}

%Our contributions are three-fold: (1) we identify and validate a low-dimensional multi-view logical subspace in the residual stream where paired NL and symbolic proofs are strongly aligned; (2) we introduce a simple, training-free steering method that amplifies projections onto this subspace during CoT generation, yielding up to 11 absolute accuracy points of improvement across five open-weight LLMs and multiple logical benchmarks, with transfer to unseen datasets; and (3) we show that this subspace encodes interpretable logical roles and that CoT projection energy correlates with correctness, making it both a practical steering handle and a window into how LLMs internally represent logical structure.

\begin{figure*}[t]
    \centering
    \includegraphics[width=0.9\linewidth]{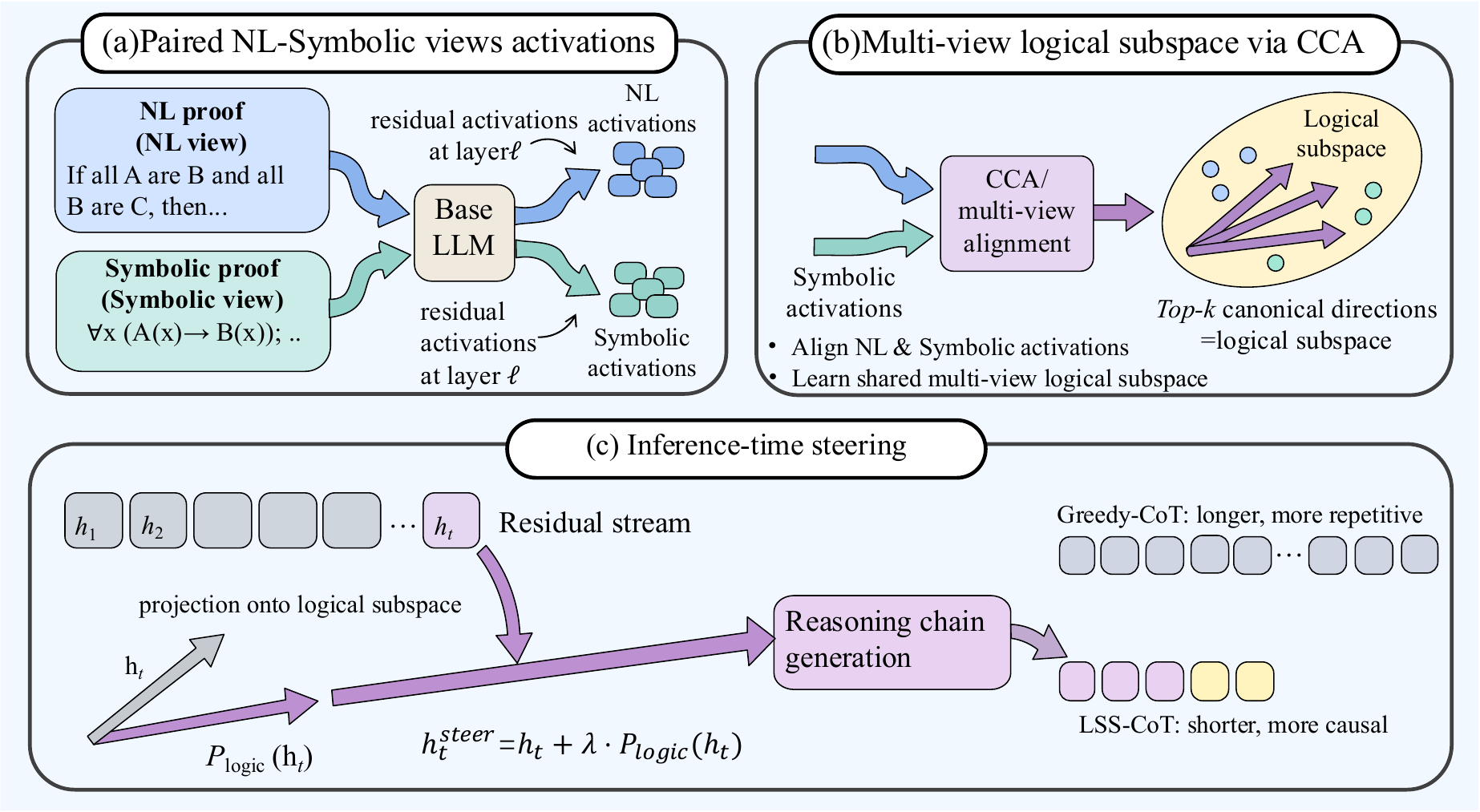}
    \caption{Overview of our logical subspace steering (LSS) method.}
    \label{fig:lss-overview}
\end{figure*}

\section{Background and Problem Setup}

\subsection{Model and Residual Stream}

We consider decoder-only language models with $L$ Transformer layers and residual dimension $D$.
Given an input token sequence $x_{1:T}$, let $h^{(\ell)}_t \in \mathbb{R}^D$ denote the residual activation at layer $\ell$ and position $t$.
Following prior work on activation-level interventions~\cite{zou2025representationengineeringtopdownapproach,turner2024steeringlanguagemodelsactivation}, we treat the residual stream $\{h^{(\ell)}_t\}$ as the main object of our analysis and steering.

\subsection{Multi-View Reasoning Data}
\label{sec:background}
Each reasoning instance $i$ consists of facts and rules, a query, a truth label, and two proof-style reasoning trajectory of the same derivation: a \emph{natural-language proof} $P_i^{\mathrm{NL}}$ written as a step-by-step explanation, and a \emph{symbolic proof} $P_i^{\mathrm{Sym}}$ given as a sequence of rule applications or logical clauses.

For each view $v \in \{\mathrm{NL}, \mathrm{Sym}\}$, we concatenate the context, query, and proof in that view into a single input and run the frozen LLM in teacher forcing, collecting residual activations $h^{(\ell)}_{i,t,v} \in \mathbb{R}^D$ at every layer $\ell$ and position $t$.

Let $\mathcal{I}^{v}_i$ be the index set of proof tokens for instance $i$ in view $v$ (excluding context and question tokens).
At each layer $\ell$, we simply mean-pool residual activations over the proof tokens in each view, yielding
$ r^{(\ell)}_{i,\mathrm{NL}}$ and $ r^{(\ell)}_{i,\mathrm{Sym}}$ for instance $i$.

Stacking the pooled representations across instances yields, for each layer $\ell$, two matrices:
\[
X^{(\ell)} \in \mathbb{R}^{N \times D}, \quad
Y^{(\ell)} \in \mathbb{R}^{N \times D},
\]
whose $i$-th rows are $r^{(\ell)}_{i,\mathrm{NL}}$ and $r^{(\ell)}_{i,\mathrm{Sym}}$, respectively.
We refer to $X^{(\ell)}$ and $Y^{(\ell)}$ as the NL and symbolic views, following standard Canonical Correlation Analysis (CCA) notation \cite{raghu2017svccasingularvectorcanonical}.

\subsection{Subspaces and Projectors}

At selected layers, we will learn from the paired NL and symbolic views a
low-dimensional subspace of the residual stream for each layer $\ell$,
represented by an orthonormal basis
$U^{(\ell)} \in \mathbb{R}^{D \times k}$ and its orthogonal projector
$P^{(\ell)} = U^{(\ell)} {U^{(\ell)}}^\top$ (details in
Section~\ref{sec:method-subspace}).
We refer to the column span of $U^{(\ell)}$ as the layer-$\ell$
\emph{multi-view logical subspace}.

\subsection{Token-level projection energy}
\label{sec:energy-def}
Given a layer-$\ell$ residual activation $r \in \mathbb{R}^D$ and the
logical subspace basis $U^{(\ell)} \in \mathbb{R}^{D \times k}$
(Section~\ref{sec:method-subspace}), we define the normalized projection energy as:
\begin{equation}
\label{eq:token-energy}
E^{(\ell)}(r)
= \frac{\big\| r^\top U^{(\ell)} \big\|_2^2}{\big\|r\big\|_2^2}.
\end{equation}
We also define the contribution of the $j$-th basis direction
$u^{(\ell)}_j$ as:
\begin{equation}
\label{eq:dir-energy}
E^{(\ell)}_j(r)
= \frac{\big( r^\top u^{(\ell)}_j \big)^2}{\big\|r\big\|_2^2},
\quad
E^{(\ell)}(r) = \sum_{j=1}^k E^{(\ell)}_j(r).
\end{equation}

Here $r^\top U^{(\ell)}$ are the coordinates of $r$ in the logical subspace, and $E^{(\ell)}(r)$ is the fraction of its $\ell_2$ norm explained by that subspace. Further details are given in Appendix~\ref{app:energy-defs}.

\section{Method}
\label{sec:method}

Our goal is to exploit the paired natural-language and symbolic proofs from
Section~\ref{sec:background} to (i) learn, for selected layers, a
low-dimensional ``multi-view logical subspace'' in the residual stream and
(ii) steer the model's hidden states along this subspace during reasoning generation.

\subsection{Learning a multi-view logical subspace}
\label{sec:method-subspace}

For each layer $\ell \in \mathcal{L}_{\mathrm{sub}}$, the
construction in Section~\ref{sec:background} yields pooled
activation matrices $X^{(\ell)}$ and $Y^{(\ell)}$ where each row corresponds to one instance, viewed through its
natural-language or symbolic proof.
We then apply a PCA+CCA pipeline to learn, on the NL side, a
low-dimensional subspace that is maximally aligned with the symbolic
view.

\paragraph{Step 1: PCA for denoising and compression.}
We first apply Principal Component Analysis (PCA) separately to
$X^{(\ell)}$ and $Y^{(\ell)}$, retaining the smallest number of
components that explain a fixed fraction of variance (98\%) and
obtaining reduced representations
\[
\tilde X^{(\ell)} \in \mathbb{R}^{N \times d_X}, \qquad
\tilde Y^{(\ell)} \in \mathbb{R}^{N \times d_Y},
\]
with $d_X, d_Y \ll D$.
We column-center $\tilde X^{(\ell)}$ and $\tilde Y^{(\ell)}$ before
running CCA.

\paragraph{Step 2: Canonical Correlation Analysis.}
We then run linear Canonical Correlation Analysis (CCA) 
\cite{cca,raghu2017svccasingularvectorcanonical} on
$\tilde X^{(\ell)}$ and $\tilde Y^{(\ell)}$, for a fixed number $k$ of canonical components ($k=32$ in the main setting;
see Appendix~\ref{app:k-robustness} for a robustness study over $k$).
CCA finds directions
$\{a_j^{(\ell)}\}_{j=1}^k$ in the NL space and
$\{b_j^{(\ell)}\}_{j=1}^k$ in the symbolic space such that the
scalar projections
$\tilde X^{(\ell)} a_j^{(\ell)}$ and
$\tilde Y^{(\ell)} b_j^{(\ell)}$
have maximal Pearson correlation.

Stacking these directions gives projection matrices
$A^{(\ell)} \in \mathbb{R}^{d_X \times k}$ and
$B^{(\ell)} \in \mathbb{R}^{d_Y \times k}$.

\paragraph{Step 3: Back-projection and orthonormal basis.}
Let $V_X^{(\ell)} \in \mathbb{R}^{D \times d_X}$ be the PCA
loading matrix for $X^{(\ell)}$.
We map the NL-side canonical directions back into the original
residual space via
\[
W^{(\ell)} = V_X^{(\ell)} A^{(\ell)} \in \mathbb{R}^{D \times k},
\]
whose columns span $k$ directions in the layer-$\ell$ residual stream.
To obtain an orthonormal basis, we apply a QR decomposition
\[
W^{(\ell)} = Q^{(\ell)} R^{(\ell)},
\]
We set $U^{(\ell)} = Q^{(\ell)} \in \mathbb{R}^{D \times k}$, which forms an orthonormal basis for the subspace, and define the corresponding projector $P^{(\ell)} = U^{(\ell)} {U^{(\ell)}}^\top$.

\subsection{Inference-time steering along the subspace}
\label{sec:method-steering}

Given the learned projectors $\{P^{(\ell)}\}$, we modify the forward pass
during Chain-of-Thought (CoT) generation without changing any model parameters.

\paragraph{Steering protocol.}
We follow a standard CoT prompting setup \citep{wei2022cot,kojima2023largelanguagemodelszeroshot}: given an input, we prepend a CoT-style instruction and let the model generate a natural-language proof and answer, intervening only on the residual stream.

Let $h^{(\ell)}_{t} \in \mathbb{R}^D$ denote the residual vector at
layer $\ell$ and time step $t$ in the forward pass.
For a chosen steering layer $\ell^\star$ and a scalar steering strength
$\lambda$, we replace $h^{(\ell^\star)}_{t}$ by
\begin{equation}
\label{eq:steer}
\tilde h^{(\ell^\star)}_{t}
= h^{(\ell^\star)}_{t}
+ \lambda \,
  \frac{P^{(\ell^\star)} h^{(\ell^\star)}_{t}}
       {\big\|P^{(\ell^\star)} h^{(\ell^\star)}_{t}\big\|_2}
  \, \big\|h^{(\ell^\star)}_{t}\big\|_2.
\end{equation}
In other words, we add a perturbation in the direction of the projection onto the multi-view logical subspace with magnitude $\lambda \|h^{(\ell^\star)}_{t}\|_2$.

In practice, we normalize
$P^{(\ell^\star)} h^{(\ell^\star)}_{t}$ with an $\varepsilon$
added to the denominator for numerical stability.

For each model--benchmark pair we use a single steering layer $\ell^\star$
and a single $\lambda$, both selected on a held-out development set
(see Appendix~\ref{app:exp-details} for the hyperparameter selection
procedure), and apply Eq.~\eqref{eq:steer} only at $\ell^\star$ for tokens
in the generated CoT; the encoding of the input context and question uses the original forward pass. This yields a training-free inference-time method that requires only a one-off subspace estimation on gold proofs and a light matrix–vector multiplication per token at the steering layer.

\section{Experiments}
\label{sec:experiments}

\subsection{Experimental Setup}

\paragraph{Benchmarks.}
We evaluate on three logical reasoning benchmarks:
\textbf{FOLIO}~\citep{folio}, a logical entailment dataset with aligned
natural-language stories and first-order logic (FOL) formalizations;
\textbf{PrOntoQA}~\citep{prontoqa}, a synthetic multi-hop ontology QA benchmark;
and \textbf{ProofWriter}~\citep{proofwriter}, which provides multi-hop
entailment questions paired with natural-language proofs under both
open-world (OWA) and closed-world (CWA) semantics.
On PrOntoQA and ProofWriter, we learn multi-view subspaces from the paired natural-language and symbolic proofs; on FOLIO, which lacks gold proofs, we instead use the aligned natural-language story and first-order logic (FOL) formalization as the two-view reasoning data.
For all benchmarks, we use splits derived from the released data or official
generation code, reserving a small development subset for hyperparameter tuning;
the exact splits and construction details are given in
Appendix~\ref{app:datasets}.
\paragraph{Models.}
We evaluate our method on five open-weight, decoder-only LLMs:
Meta-Llama-3.1-8B-Instruct, Llama-3.2-3B-Instruct~\citep{llama3},
Llama-2-13B-Chat~\citep{llama2},
Gemma-2-9B-IT~\citep{gemma2},
and Phi-3-Mini-4K-Instruct~\citep{phi3}.
We use publicly released checkpoints and tokenizers without any additional
fine-tuning, and estimate multi-view logical subspaces separately for each
model–benchmark pair. Full model and decoding details are provided in
Appendix~\ref{app:models}.

\begin{table*}[t]
\centering
\footnotesize
\setlength{\tabcolsep}{5pt}
{\renewcommand{\arraystretch}{0.4}
\scalebox{1.0}{
\begin{tabular}{llcccc}
\toprule
\multirow{2}{*}{Model} & \multirow{2}{*}{Setting} &
\multicolumn{4}{c}{Accuracy (\%)} \\
\cmidrule(lr){3-6}
& & FOLIO & PrOntoQA (5-hop) & PW-CWA (3-hop) & PW-OWA (3-hop) \\
\midrule
\multirow{5}{*}{Llama-3.1-8B Instruct}
& Greedy-CoT  & 51.7 & 70.6 & 51.4 & 50.7 \\
& 3-shot-CoT  & \underline{60.6} & 72.4 & \textbf{66.4} & \textbf{63.7} \\
& SC-3        & 58.1 & \textbf{79.0} & 47.2 & 53.3 \\
\rowcolor{steerbg}
& \textbf{\ourmethod{}} & \textbf{61.1} & \underline{75.4} & \underline{55.6} & \underline{55.3} \\
& $\Delta$ (LSS--Greedy) & $\uparrow 9.4$ & $\uparrow 4.8$ & $\uparrow 4.2$ & $\uparrow 4.6$ \\
\midrule
\multirow{5}{*}{LLaMA-3.2-3B Instruct}
& Greedy-CoT  & \underline{51.2} & 50.0 & 46.6 & 37.1 \\
& 3-shot-CoT  & 42.4 & \textbf{61.4} & \textbf{56.8} & \textbf{51.5} \\
& SC-3        & 49.3 & 49.4 & 46.8 & 37.1 \\
\rowcolor{steerbg}
& \textbf{\ourmethod{}} & \textbf{53.7} & \underline{53.4} & \underline{49.8} & \underline{38.7} \\
& $\Delta$ (LSS--Greedy) & $\uparrow 2.5$ & $\uparrow 3.4$ & $\uparrow 3.2$ & $\uparrow 1.6$ \\
\midrule
\multirow{5}{*}{LLaMA-2-13B Chat}
& Greedy-CoT  & 42.9 & 48.6 & 54.6 & 39.1 \\
& 3-shot-CoT  & \textbf{51.2} & 51.2 & \textbf{58.0} & \textbf{42.9} \\
& SC-3        & \underline{46.8} & \textbf{55.4} & 52.6 & 42.5 \\
\rowcolor{steerbg}
& \textbf{\ourmethod{}} & 45.8 & \underline{55.0} & \underline{56.6} & \textbf{42.9} \\
& $\Delta$ (LSS--Greedy) & $\uparrow 2.9$ & $\uparrow 6.4$ & $\uparrow 2.0$ & $\uparrow 3.8$ \\
\midrule
\multirow{5}{*}{Gemma-2-9B It}
& Greedy-CoT  & 58.1 & 87.4 & 71.4 & 75.0 \\
& 3-shot-CoT  & \textbf{68.0} & 82.8 & \underline{72.8} & 71.9 \\
& SC-3        & 63.1 & \underline{90.0} & \underline{72.8} & \underline{75.6} \\
\rowcolor{steerbg}
& \textbf{\ourmethod{}} & \underline{65.5} & \textbf{90.2} & \textbf{73.8} & \textbf{76.6} \\
& $\Delta$ (LSS--Greedy) & $\uparrow 7.4$ & $\uparrow 2.8$ & $\uparrow 2.4$ & $\uparrow 1.6$ \\
\midrule
\multirow{5}{*}{Phi-3-mini-4k Instruct}
& Greedy-CoT  & 62.6 & 59.6 & 59.6 & 51.1 \\
& 3-shot-CoT  & \textbf{68.0} & 63.8 & \textbf{69.2} & \textbf{73.9} \\
& SC-3        & \underline{67.0} & \underline{70.2} & \underline{66.4} & \underline{59.3} \\
\rowcolor{steerbg}
& \textbf{\ourmethod{}} & 66.0 & \textbf{70.6} & 63.4 & 56.1 \\
& $\Delta$ (LSS--Greedy) & $\uparrow 3.4$ & $\uparrow 11.0$ & $\uparrow 3.8$ & $\uparrow 5.0$ \\

\bottomrule
\end{tabular}
}
}
\caption{
Main results on four logical reasoning benchmarks.
Greedy-CoT, \ourmethod{}, SC-3, and 3-shot-CoT are defined in Section~\ref{sec:experiments}.
For each model, the last row shows LSS--Greedy absolute improvement.
Best score per model–benchmark pair is in \textbf{bold}, second-best is \underline{underlined}.}
\label{tab:main-results}
\end{table*}

\paragraph{Baselines.}
For each model–benchmark pair we compare with the following decoding variants:
\begin{itemize}
\item \textbf{Greedy-CoT}: zero-shot CoT prompting
\citep{wei2022cot} with greedy decoding and no activation intervention.
    \item \textbf{3-shot-CoT}: few-shot CoT prompting
\citep{brown2020languagemodelsfewshotlearners,wei2022cot} with three
in-context exemplars (Appendix~\ref{app:datasets}).
    \item \textbf{SC-3} (self-consistency): CoT prompting with three
sampled reasoning paths per instance, following
\citet{wang2023selfconsistencyimproveschainthought}; implementation details are given in Appendix~\ref{app:exp-details}.

\end{itemize}

\paragraph{Experiment Setting.} Our proposed \textbf{\ourmethod{}} method applies
single-layer steering along the learned multi-view logical subspace during
CoT generation (Section~\ref{sec:method-steering}; tuning details in
Appendix~\ref{app:exp-details}). Across all benchmarks we use a single
task-specific CoT prompt per task (Appendix~\ref{app:prompts}) and, unless
otherwise noted, greedy decoding with a budget of 1024 new tokens, treating
cases with no answer within this budget as incorrect. Inference overhead is negligible in practice: on our {PrOntoQA} setup with Llama-3.1-8B-Instruct, throughput is 179 tok/s without
steering and 176 tok/s with steering under the same experimental setting.

\subsection{Main Results}
\label{sec:main-results}

Table~\ref{tab:main-results} presents our main results across
benchmarks and models.
Overall, steering along the multi-view logical subspace consistently
improves Chain-of-Thought (CoT) accuracy over the Greedy-CoT baseline
and often matches or surpasses much more expensive self-consistency
decoding.

\paragraph{Gains over Greedy CoT.}
Across all model--benchmark pairs, \textbf{\ourmethod{}} improves over
\textbf{Greedy-CoT} by between 1.6 and 11 absolute accuracy points.
The largest gains appear on PrOntoQA for {Phi-3-Mini} ($+11.0$)
and on FOLIO for {Llama-3.1-8B} ($+9.4$), where the base
Greedy-CoT performance leaves substantial headroom.
Even on stronger configurations such as {Gemma-2-9B} on
PrOntoQA and ProofWriter, steering still yields non-trivial
improvements (roughly $+2$--$+3$ points), showing that the learned
multi-view subspace adds value beyond what pretrained models already
achieve with standard CoT prompting. LSS also remains effective on a reasoning-specialized model.
On {Qwen3-4B} evaluated on {PrOntoQA}, accuracy improves
from 87.2 without steering to 93.2 with LSS (+6.0), indicating that the
learned multi-view subspace remains beneficial even when the base model
already exhibits strong reasoning performance (Appendix~\ref{app:qwen3}).

\paragraph{Comparison to SC-3.}
Self-consistency with $k=3$ (\textbf{SC-3}) provides a strong
test-time scaling baseline, but incurs a $3\times$ increase in
inference cost.
On several settings, \textbf{\ourmethod{}} with a single decoding
trajectory matches or even outperforms SC-3.
For example, on FOLIO with {Llama-3.1-8B}, our method improves
accuracy from 51.7 to 61.1, whereas SC-3 reaches only 58.1.
On PrOntoQA, {Gemma-2-9B} and {Phi-3-Mini} achieve
comparable or slightly higher accuracy under \ourmethod{} than under
SC-3, despite using only one CoT sampling process instead of three.
These results suggest that activation-level steering can recover much
of the benefit of test-time ensembling at a fraction of the compute
cost.

On smaller models such as {Llama-3.2-3B}, SC-3 can even degrade
performance (e.g., on FOLIO and PrOntoQA) relative to Greedy-CoT. One possible explanation is that its sampled reasoning paths exhibit high variance and
introduce additional noise.

In contrast, \ourmethod{} consistently delivers modest but reliable
improvements on the same model, indicating that manipulating internal
activations can serve as a \textbf{stabilizer} for weaker architectures:
rather than adding more samples, it nudges hidden states toward
reasoning-relevant configurations in a controlled way.

\paragraph{Comparison to few-shot CoT.}
We also compare against \textbf{3-shot-CoT}, which augments the prompt
with three in-context examples.
On FOLIO, \ourmethod{} \emph{outperforms or matches} 3-shot-CoT on the
LLaMA-3 models while using no in-context examples and thus incurring zero
additional prompt length.
For example, on {Llama-3.1-8B} and {Llama-3.2-3B},
\ourmethod{} achieves higher accuracy than 3-shot-CoT despite operating
in a purely zero-shot setting, whereas 3-shot-CoT can even degrade
performance on the smaller {Llama-3.2-3B}.
This highlights a complementary form of \emph{context efficiency}:
rather than relying on longer prompts or manually curated exemplars,
our method improves reasoning purely via a lightweight intervention on
the residual stream.

Taken together, these results show that steering along the learned
multi-view logical subspace yields consistent gains over Greedy-CoT,
is competitive with self-consistency at far lower compute cost, and
can outperform few-shot CoT without using additional context tokens.

More broadly, they point to a third, complementary route for
enhancing reasoning in LLMs: instead of scaling context length or
sampling budget, we can directly \emph{align internal representations}
via activation-level steering, effectively unlocking latent logical
capabilities that are underutilized by standard decoding.

\begin{table}[t]
\centering
\small
\setlength{\tabcolsep}{6pt}
\begin{tabular}{@{}lc@{}}
\toprule
\textbf{Method} & \textbf{Accuracy (\%)} \\
\midrule
Greedy-CoT & 70.6 \\
3-shot-CoT & 72.4 \\
3-shot-CoT + LSS & 74.6 \\
SC-3 & 79.0 \\
SC-3 + LSS & \textbf{81.0} \\
\bottomrule
\end{tabular}
\caption{Combining LSS with inference schemes (few-shot CoT and self-consistency) on {PrOntoQA} (5-hop) using {Llama-3.1-8B-Instruct}.}
\label{tab:lss-stacking}
\end{table}

\subsection{Compatibility with Inference Schemes}
Beyond comparing \ourmethod{} against the decoding baselines such as few-shot CoT and self consistency, we also
test whether the same learned logical subspace can be stacked on top of them.
On {PrOntoQA} (5-hop) with {Llama-3.1-8B-Instruct}, we reuse
the exact same subspace, steering layer, and strength $\lambda$ selected for
\ourmethod{} in the main experiment, with no additional retuning.
As shown in Table~\ref{tab:lss-stacking}, adding LSS on top of 3-shot CoT
improves accuracy from 72.4 to 74.6 (+2.2), and adding LSS on top of SC-3
improves accuracy from 79.0 to 81.0 (+2.0), with SC-3+LSS performing best
overall. These results show that LSS is not only a lightweight alternative to
more expensive inference schemes, but also a complementary intervention that
can further improve them.
\begin{figure}[t]
    \centering
    \includegraphics[width=0.9\linewidth]{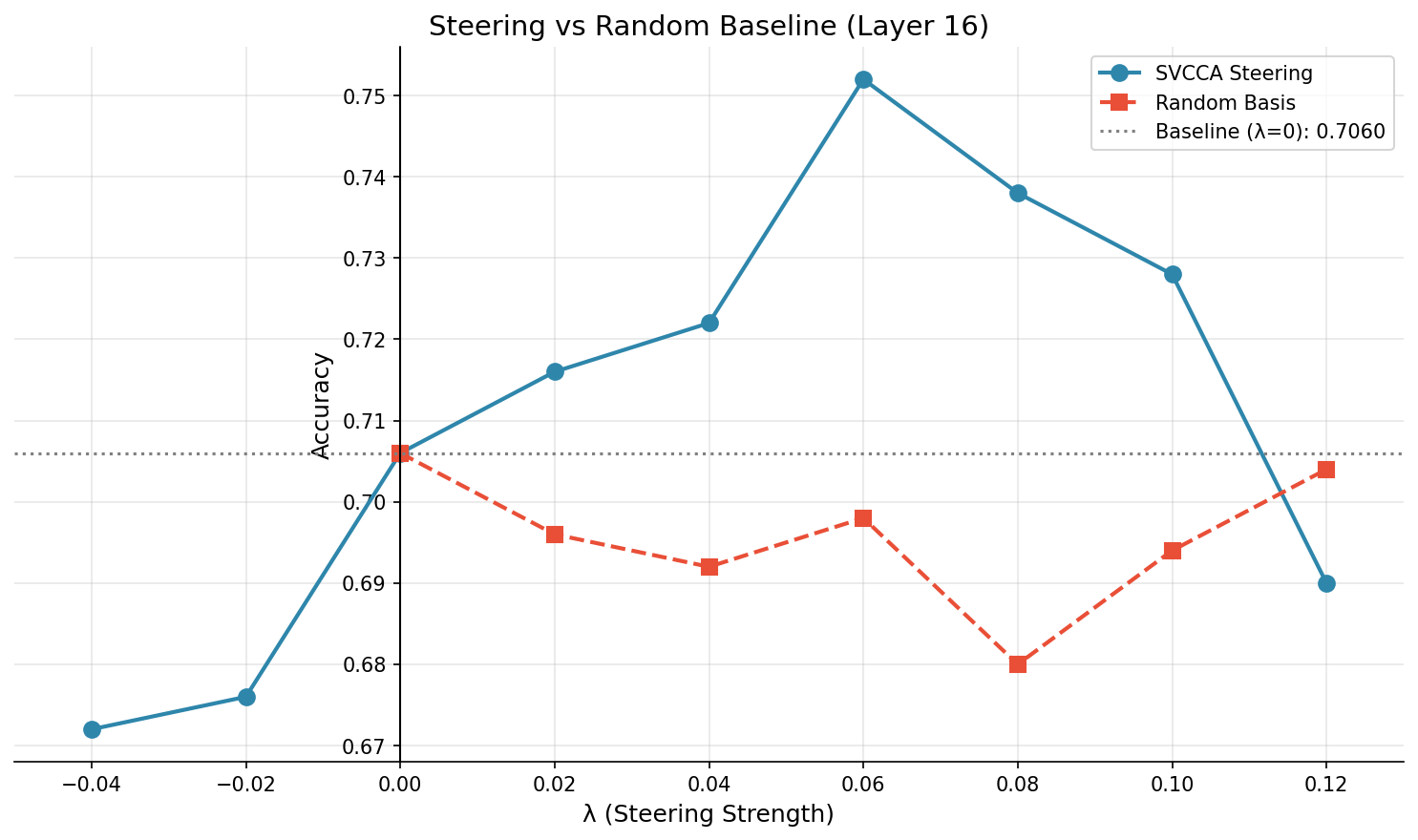}%
    \caption{
    Sensitivity to steering strength.
    Accuracy on PrOntoQA ({Llama-3.1-8B}) as a function of steering
    coefficient $\lambda$ for our logical subspace direction (blue) vs.\ random
    orthogonal directions (red).
    }
    \label{fig:sensitivity}
\end{figure}
\subsection{Sensitivity to the Steering Direction}
\label{sec:sensitivity}
We study how performance varies with the steering coefficient $\lambda$ and
direction on PrOntoQA with {Llama-3.1-8B}.
At the chosen layer, we compare steering along our logical subspace direction
to steering along random orthogonal directions in the same residual space
(Fig.~\ref{fig:sensitivity}; further details are given in
Appendix~\ref{sec:appendix_steering_robustness}).

Steering along random directions fails to yield systematic gains and often
degrades accuracy across $\lambda$.
By contrast, steering along the logical subspace direction shows a clear
asymmetry: moderate positive $\lambda$ improves performance, while negative
$\lambda$ causes sharp drops.
This pattern supports the view that the extracted direction is
\emph{task-positive}, rather than a generic rescaling of activations.

\begin{table}[t]
\centering
\small
\setlength{\tabcolsep}{4pt}
\begin{tabular}{@{}llccr@{}}
\toprule
\multirow{2}{*}{Model} & \multirow{2}{*}{Setting} & \multicolumn{3}{c}{Accuracy (\%)} \\
\cmidrule(lr){3-5}
& & Greedy & LSS & $\Delta$ \\
\midrule
\multirow{2}{*}{{Llama-3.1-8B}}
& NLI & 51.8 & \textbf{56.2} & $\uparrow\,4.4$ \\
& MCR & 48.8 & \textbf{51.8} & $\uparrow\,3.0$ \\
\midrule
\multirow{2}{*}{{Phi-3-Mini}}
& NLI & 52.6 & \textbf{56.2} & $\uparrow\,3.6$ \\
& MCR & 51.8 & \textbf{54.4} & $\uparrow\,2.6$ \\
\bottomrule
\end{tabular}

\caption{Cross-dataset generalization on LogiQA2.0}
\label{tab:generalization}
\end{table}

\subsection{Generalization}
\label{sec:generalization}

We test whether the multi-view logical subspace captures abstract reasoning
patterns that transfer beyond the synthetic data it was learned on.

Concretely, we keep the PrOntoQA-trained subspace $P^{(\ell)}$ fixed
(Section~\ref{sec:encoding}) and use it to steer models on the LogiQA 2.0 dataset~\citep{logiqa}, evaluating both its natural language inference (NLI) and multiple-choice
reading (MCR) formats.
The subspace is learned only from synthetic rule-based PrOntoQA proofs and will not be updated on LogiQA; full setup details, including sample sizes, are given
in Appendix~\ref{app:gen-setup}.

% ==========================================
% TABLE: GENERALIZATION RESULTS (LogiQA)
% ==========================================

Table~\ref{tab:generalization} shows that this PrOntoQA-derived subspace
transfers effectively to LogiQA~2.0: steering yields consistent accuracy gains
across models and both evaluation formats (e.g., \textbf{+3.0} points on
{Llama-3.1-8B} MCR). We observe a similar transfer effect on
{ReClor} dataset \citep{yu2020reclor}, where reusing the same fixed PrOntoQA-derived subspace
improves {Llama-3.1-8B} from 53.4 to 56.6 (+3.2), without retraining
the subspace.

Despite the domain shift from clean synthetic proofs to noisy exam-style
passages and multiple-choice logical reading comprehension, these improvements
suggest that the multi-view subspace encodes general logical mechanisms, such
as rule following and entailment tracking, rather than overfitting to the
surface form of PrOntoQA.

We further conduct a non-logic sanity check, by evaluating the effect of steering on a non-logic commonsense reasoning dataset HellaSwag \citep{zellers2019hellaswag}. Specifically, we steer Llama-3.1-8B using the same PrOntoQA-derived subspace, layer, and $\lambda$ as in the main experiment. Accuracy changes only from 0.652 to 0.650, indicating negligible degradation on the non-logic reasoning benchmark.

%As a non-logic sanity check, we also evaluate steering on {HellaSwag} dataset \citep{zellers2019hellaswag} which is a non-logic commonsense reasoning benchmark using the same {PrOntoQA}-derived subspace, layer, and $\lambda$ as in the main {Llama-3.1-8B} experiment. Accuracy changes only from 0.652 to 0.650, indicating negligible degradation on the non-logic task.

\section{Analysis}
\label{sec:analysis}

We next analyze the learned multi-view logical subspaces and the effect
of steering on model behavior. Our goal is to better understand
(i) what these logical subspaces encode,
(ii) how NL-symbolic alignment evolves within model,
(iii) how the logical subspace relate to reasoning success, and
(iv) how steering reshapes the structure of CoT generation.
\begin{figure}[t]
    \centering
    \includegraphics[width=\linewidth]{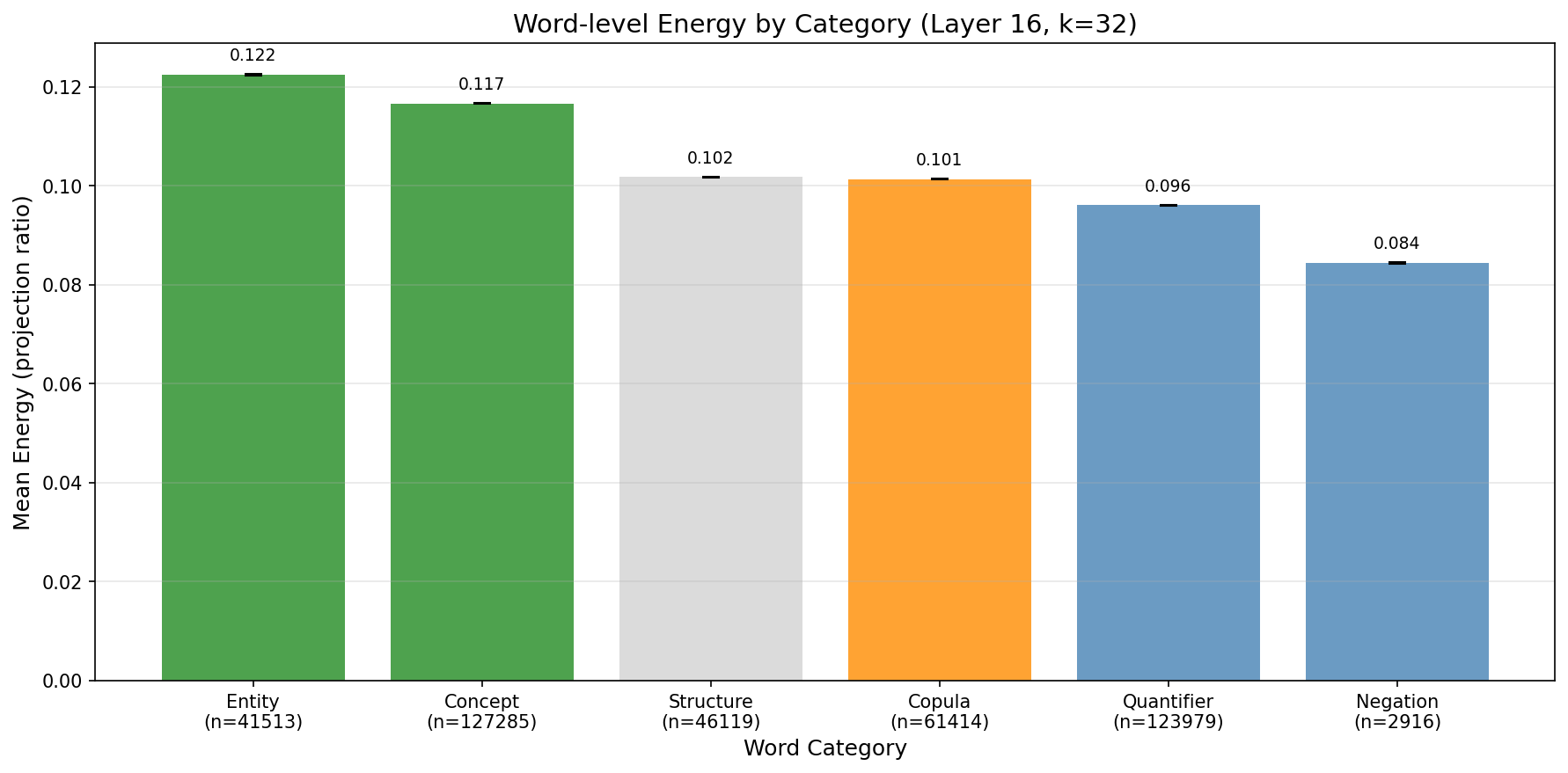}
\caption{\textbf{Global Energy Distribution.} Average projection energy by category ({Llama-3.1-8B}, Layer 16).}
    \label{fig:category-energy}
\end{figure}
\subsection{What does the multi-view subspace encode?}
\label{sec:encoding}

To understand what the learned multi-view logical subspaces capture, we use
the token-level projection energy $E^{(\ell)}(r)$ defined in
Eq.~\ref{eq:token-energy}, which measures how much of a token’s representation lies inside the logical subspace, and its per-direction decomposition (Eq.~\ref{eq:dir-energy}).
We apply this analysis to a synthetic rule-based corpus generated with the
official PrOntoQA code; details in Appendix~\ref{app:energy-data}.

\begin{figure}[t]
    \centering
    \includegraphics[width=\linewidth]{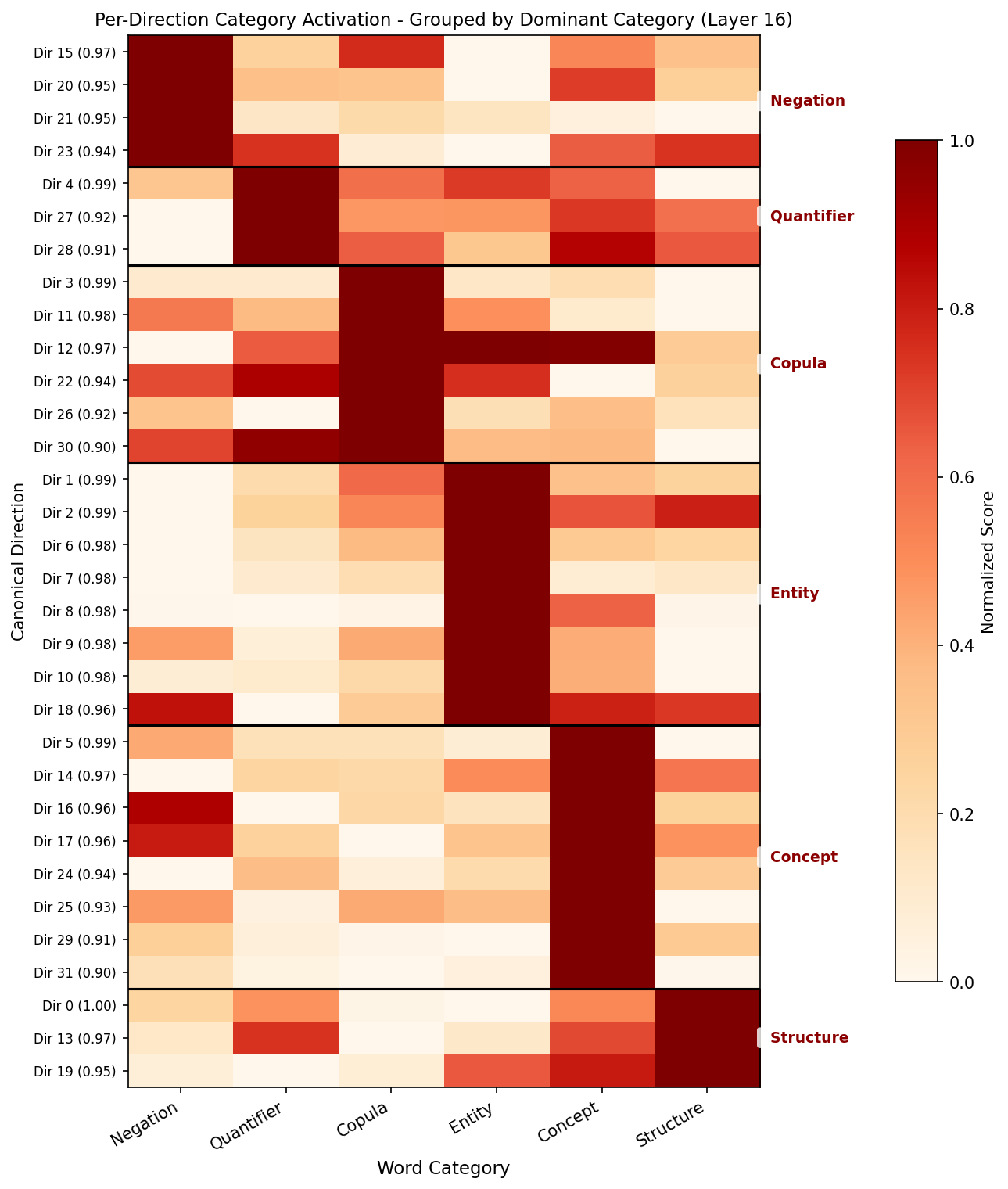} 
\caption{
\textbf{Heatmap of directional selectivity.}
Normalized activations for {Llama-3.1-8B} (Layer 16), with directions \textbf{sorted by dominant category}.
}
    \label{fig:per-dir-heatmap}
\end{figure}

At a global level, we ask which token categories carry most of the subspace energy.  We aggregate $E^{(\ell)}(r)$ over tokens of each
category within a proof and then average across instances.
Across both {Llama-3.1-8B} and {Phi-3-Mini}, entity and concept
tokens have the highest mean energy, followed by structure and copula tokens,
while negations and quantifiers carry lower but still clearly non-zero energy.
In other words, the shared subspace allocates most of its mass to “who”
the proof is about and “what” properties or relations hold, while still
systematically responding to logical operators rather than treating them
as noise.
Similar patterns appear across several middle and upper layers, suggesting
that this focus on core predicate content is a stable property of the
subspace
(see Fig.~\ref{fig:category-energy} and Appendix~\ref{app:energy-defs}).

At a finer-grained level, we ask whether individual canonical directions
$u^{(\ell)}_j$ correspond to specific logical roles. We compute
their average normalized contribution to each word category and visualize
the resulting matrix.
Directions cluster into interpretable groups: some are dominated by entities
and concepts, others by quantifiers or negation, and others by structural
markers such as connectives and conclusion phrases
(Fig.~\ref{fig:per-dir-heatmap}).
This “role-wise’’ organization is consistent between layers and models,
indicating that the multi-view logical subspace is not an arbitrary slice of
the residual stream but a structured space whose basis directions specialize
in different aspects of the proof.
Additional visualizations are provided in Appendix~\ref{app:energy-viz}.

\subsection{How does the alignment between language and logic evolve across layers?}
\label{sec:layer-dynamics}
We next ask how strongly the natural-language and symbolic views are aligned across layers. Following prior work on representation similarity \citep{raghu2017svccasingularvectorcanonical},
we compute, for each layer $\ell$, the mean canonical correlation
$\bar{\rho}^{(\ell)}$ between the pooled NL and symbolic activations on PrOntoQA
(details in Appendix~\ref{app:analysis-setup}).
Intuitively, $\bar{\rho}^{(\ell)}$ ranges from 0 (no shared structure) to 1
(perfect alignment).

\begin{figure}[t]
    \centering
    \includegraphics[width=0.9\linewidth]{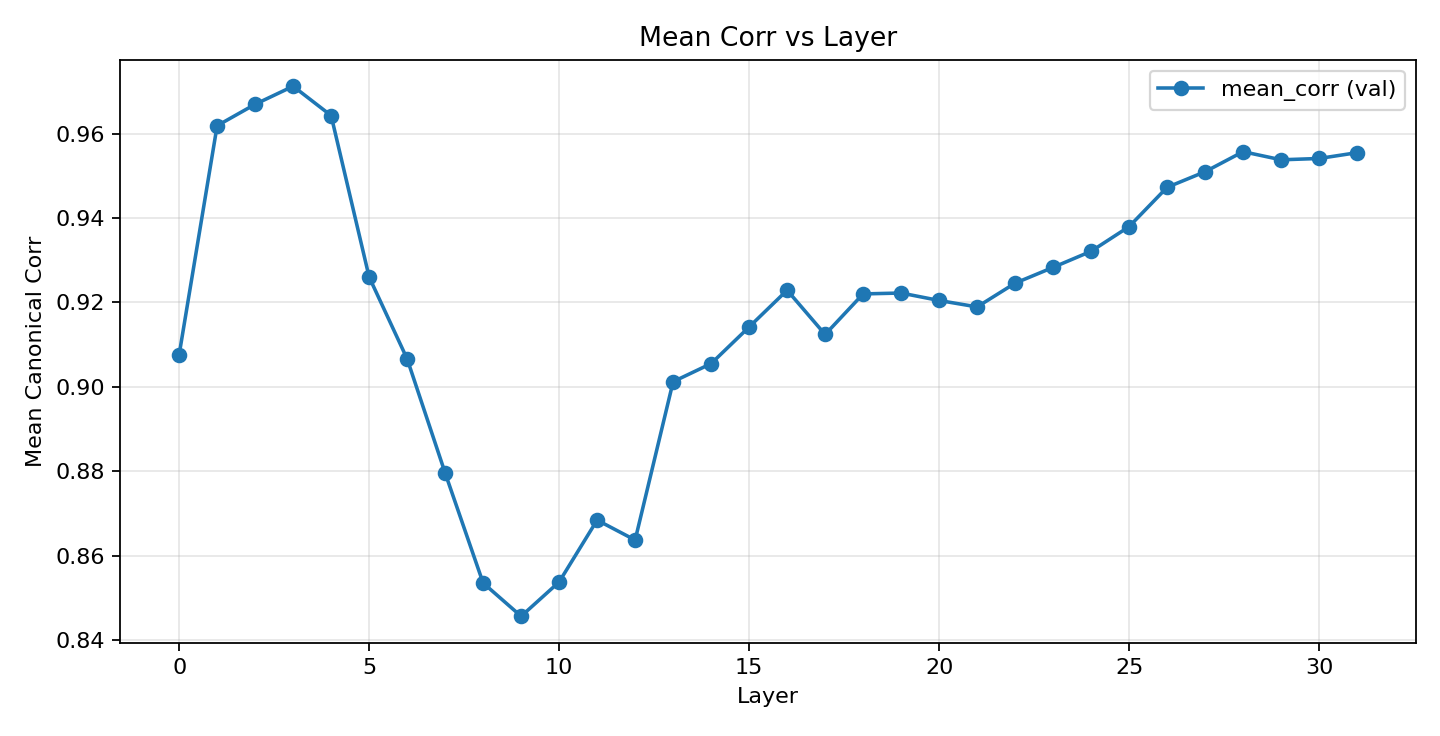}
    \caption{
    Layer-wise mean canonical correlation between language and logic ({Llama-3.1-8B}).}
    
    \label{fig:layer-corr}
\end{figure}
Figure~\ref{fig:layer-corr} shows the resulting trajectory for {Llama-3.1-8B}.
Across all evaluated models we observe a robust \textbf{``high--low--high''}
pattern: alignment is high in the early layers, drops in the middle layers,
and rises again in upper layers (full curves in
Appendix~\ref{app:layer-curves}).
This suggests that the logical subspace is \emph{dynamic}: alignment is weakened in the middle layers, where the model appears to encode information into more task-specific representations, and then strengthened again near the top.

\subsection{Can the logical subspace tell when a reasoning chain is correct?}
\label{sec:energy-signal}
We now ask whether activations that lie more strongly in the logical
subspace are also associated with correct CoTs.
\begin{figure}[t]
    \centering
    \includegraphics[width=0.95\linewidth]{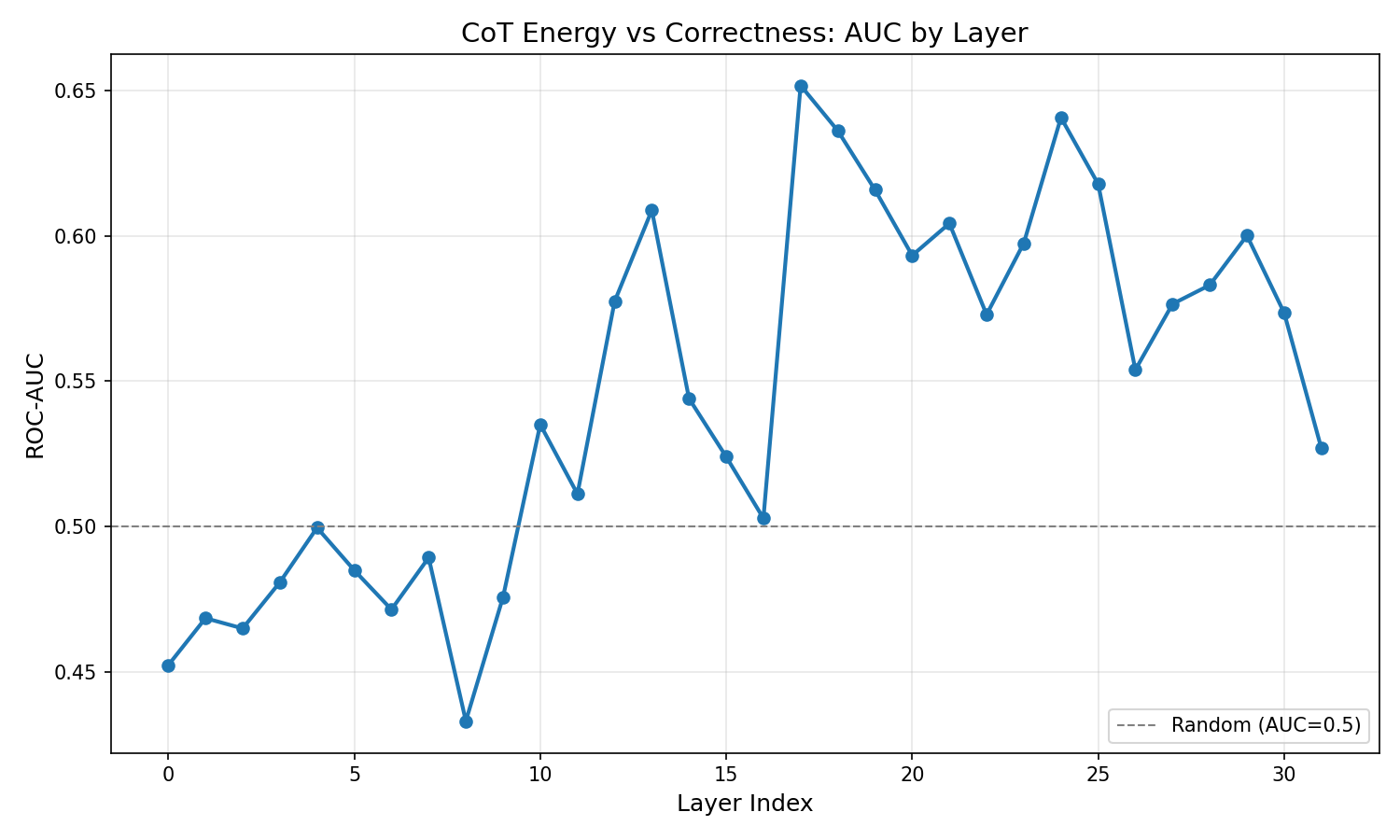}
    \caption{
    Layer-wise ROC-AUC of projection-energy-based correctness prediction
    on PrOntoQA for {Llama-3.1-8B}.
    }
    \label{fig:energy-auc-layer}
\end{figure}
For {Llama-3.1-8B} on PrOntoQA, we reuse the layer-wise projectors
$P^{(\ell)}$ from the steering setup and score Greedy-CoT traces, for each
layer, by their mean projection energy over CoT tokens (see
Appendix~\ref{app:analysis-setup} for details).
We treat this mean energy as a scalar score for each chain and, for
each layer, compute the ROC-AUC of this score for distinguishing correct
from incorrect answers (Fig.~\ref{fig:energy-auc-layer}). An AUC of 0.5 corresponds to random guessing and 1.0 to perfect discrimination.

Across most layers, projection energy yields AUCs clearly above 0.5,
indicating that correct reasoning paths tend to align more strongly with
the learned logical subspace.
The signal is strongest in middle-to-late layers (peaking around
AUC~$\approx 0.65$), consistent with the alignment dynamics in
Section~\ref{sec:layer-dynamics}: after the mid-layer ``decoupling''
phase, deeper layers again bring language and logic into closer alignment,
so projection energy provides a non-trivial signal of CoT correctness.

\subsection{How does steering change the model's reasoning behavior?}
\label{sec:behavioral_analysis}

We analyze how steering changes CoT behavior on PrOntoQA for {Llama-3.1-8B}, using the same setup as in Section~\ref{sec:experiments} and computing statistics over the CoT span only (details in Appendix~\ref{app:analysis-setup}).

Steering noticeably reshapes both length and style.
CoTs that are already correct under Greedy-CoT become slightly longer (on average \textbf{+2.11} steps), whereas those incorrect ones become much shorter (on average \textbf{$-9.83$} steps), suggesting that steering prunes unproductive loops.
Lexical counts of key reasoning markers (Appendix~\ref{app:style-lexicons}) show a similar shift: ``reasoning verbs'' such as \emph{think}, \emph{know}, and \emph{assume} decrease by \textbf{16.7\%}, while logical connectives increase by \textbf{4.8\%}, with \emph{since} and \emph{so} showing the largest gains (\textbf{+7.0\%} and \textbf{+18.3\%}).
Overall, \ourmethod{} uses fewer conversational fillers and more explicit causal links (e.g., ``since A, so B''), yielding shorter and more deductive chains.

These aggregate trends are mirrored in individual examples.
Table~\ref{tab:case-study-mini} shows an abridged PrOntoQA instance where Greedy-CoT enters an overlong self-checking loop and predicts an incorrect label, whereas \ourmethod{} follows a short causal chain and stops early with the correct answer.
\begin{table}[t]
\centering
\small
\setlength{\tabcolsep}{4pt}
\renewcommand{\arraystretch}{1.15}
\begin{tabular}{@{}p{0.22\linewidth}p{0.74\linewidth}@{}}
\toprule
\textbf{Method} & \textbf{Reasoning chain} \\
\midrule
\textbf{Greedy-CoT} &
\textit{(After reaching the key fact, continues with irrelevant checks.)}
\textcolor{badred}{``we cannot conclude anything about Polly''}
\textcolor{badred}{``we cannot conclude anything''}
\textit{(repeats for many more steps)} $\Rightarrow$ \textbf{wrong label}. \\
\textbf{\ourmethod{}} &
\textit{(Goal-directed, early stop.)}
\textcolor{goodgreen}{``Since Polly is a shumpus and shumpuses are not snowy, therefore Polly is not snowy.''}
$\Rightarrow$ \textbf{correct label}. \\
\bottomrule
\end{tabular}
\caption{PrOntoQA case study (abridged).}
\label{tab:case-study-mini}
\end{table}

\section{Related Work}

Chain-of-Thought (CoT) prompting improves multi-step reasoning in large language
models by eliciting intermediate reasoning steps rather than direct answers.
Subsequent work develops stronger test-time inference schemes, such as
self-consistency and other compute-scaled decoding strategies, to further
improve reasoning robustness
\citep{wei2022cot,kojima2023largelanguagemodelszeroshot,wang2023selfconsistencyimproveschainthought,long2023largelanguagemodelguided,zhou2023leasttomostpromptingenablescomplex}.

A related line of work seeks to combine natural-language and symbolic
reasoning. One class of methods does so at inference time by translating
natural-language inputs or intermediate reasoning into symbolic forms and
invoking external provers, solvers, or verifiers
\citep{olausson-etal-2023-linc,pan-etal-2023-logic,pei-etal-2025-fover}.

Another class uses training or finetuning with natural-language and symbolic
supervision so that the model internalizes stronger alignment between the two
views during parameter learning
\citep{zhou-etal-2025-dissecting, tan-etal-2025-enhancing-logical}.
These approaches improve reasoning either through external tool use or through
additional supervision.

Activation steering provides a complementary way to improve model behavior by
intervening directly in internal representations at inference time, often using
directions learned from contrastive activations, representation engineering, or
sparse features
\citep{turner2024steeringlanguagemodelsactivation,rimsky-etal-2024-steering,li-etal-2025-feature,galichin2025icoveredbaseshere}.

Our work offers a new perspective on combining natural-language and symbolic
reasoning. Instead of relying on external symbolic tools at inference time or
using additional supervision to internalize the alignment during training, we
ask whether a shared logical subspace aligning the two views is already
internalized in a LLM. We then operationalize this perspective through
a training-free activation intervention: using paired natural-language and
symbolic proofs, we estimate a shared multi-view logical subspace and steer
CoT generation along it without changing model weights or relying on external
symbolic solvers.

\section{Conclusion}

We introduce logical subspace steering, which uses paired natural-language and symbolic proofs to estimate a multi-view logical subspace in a frozen LLM and steer Chain-of-Thought reasoning at inference time. Across multiple logical reasoning benchmarks, this LSS-CoT outperforms greedy CoT and recovers much of the benefit of more expensive self-consistency and few-shot prompting without changing model weights or relying on external provers. Analysis shows that steering yields shorter, more causal reasoning chains and interpretable logical structure, giving a representation-level handle on the model’s logical reasoning.

\section*{Limitations}

Our work focuses on logical reasoning, and the findings may not directly generalize to broader settings such as open-domain dialogue, mathematical problem solving, long-horizon agentic tasks, multilingual applications, or extremely large proprietary or specialized theorem-proving models. Evaluating whether similar multi-view logical subspaces arise, and remain amenable to steering, in these broader domains, architectures, and training regimes is an important direction for future work. In addition, we primarily evaluate reasoning quality through final-answer accuracy. Developing more fine-grained evaluations of intermediate reasoning steps, including their correctness and faithfulness, is a valuable avenue for future research.

\section*{Ethics Statement}

This work studies representation-level methods for improving logical reasoning in open-weight language models by steering internal activations along a learned multi-view logical subspace. Our experiments are conducted on publicly available benchmarks (FOLIO, PrOntoQA, ProofWriter, and LogiQA~2.0) and open-weight models within the 3B--13B parameter range, and do not involve private user data or human subjects. We release code and data with the expectation that they will be used responsibly for research and social good.

%Because logical reasoning capabilities can amplify both beneficial and harmful uses of language models, any deployment of our method should consider downstream application context. In particular, activating or strengthening logical reasoning may improve the reliability of systems used for education, scientific assistance, or formal verification, but could also inadvertently aid misuse scenarios that benefit from better planning or proof-like reasoning. We do not deploy or evaluate our method in safety-critical settings, and we recommend that future work combine logical subspace steering with appropriate safeguards and domain-specific oversight when applied beyond the controlled benchmarks considered here.

\bibliography{main}

\appendix

\section{Dataset Construction and Splits}
\label{app:datasets}
\subsection{FOLIO}

FOLIO~\citep{folio} provides 1{,}001 training instances and 203
development instances, but no public test set.
Each instance contains natural-language premises, a corresponding
first-order logic (FOL) formalization, a hypothesis, and a three-way
label (\textsc{True}, \textsc{False}, \textsc{Uncertain}).

Since FOLIO does not include gold proofs, we treat the premises plus
hypothesis as the core reasoning context, and use the ground-truth
label to form complete inputs for both views when constructing
multi-view activations.
Concretely, in the natural-language view we concatenate the premises,
hypothesis, and a final sentence of the form
\textit{``The hypothesis is [label].''};
in the symbolic view we concatenate the FOL formalization, the
hypothesis in FOL, and the same label statement in natural language.
These two inputs describe the same entailment judgement in different
forms and are fed to the frozen model in a teacher-forcing pass to
collect residual activations (Section~\ref{sec:background}).
Note that this label information is used only for subspace estimation
and is never available at test time when generating CoTs or evaluating
accuracy.

For our experiments, we use the released development set as the test set.
From the 1{,}001 training instances, we randomly sample 50 examples as a
held-out development set for hyperparameter tuning and use the remaining
951 examples to estimate the multi-view logical subspaces.
All random splits are controlled by a fixed seed for reproducibility.

\subsubsection*{FOLIO example instance}
\label{app:folio-example}

For illustration, we show a toy FOLIO-style instance and how we form
the two views used for activation extraction.

\paragraph{Raw fields (toy example).}
\begin{itemize}
  \item Premises (NL):\\
  \textit{All birds are animals. Tweety is a bird.}
  \item Premises (FOL):\\
  \textit{$\forall x\,(\mathrm{Bird}(x) \rightarrow \mathrm{Animal}(x))$\\
         $\mathrm{Bird}(\mathrm{tweety})$}
  \item Conclusion (NL):\\
  \textit{Tweety is an animal.}
  \item Conclusion (FOL):\\
  \textit{$\mathrm{Animal}(\mathrm{tweety})$}
  \item Label: \textsc{True}
\end{itemize}

\paragraph{NL-view input.}
\textit{Premises:\\
All birds are animals. Tweety is a bird.\\[0.3em]
Hypothesis:\\
Tweety is an animal.\\[0.3em]
The Hypothesis is True.}

\paragraph{Symbolic-view input.}
\textit{Premises (FOL):\\
$\forall x\,(\mathrm{Bird}(x) \rightarrow \mathrm{Animal}(x))$\\
$\mathrm{Bird}(\mathrm{tweety})$\\[0.3em]
Hypothesis (FOL):\\
$\mathrm{Animal}(\mathrm{tweety})$\\[0.3em]
The Hypothesis is True.}
\subsection{PrOntoQA}

For PrOntoQA~\citep{prontoqa}, we follow the \textsc{Fictional-5hop}
setting and use the official synthetic data generation code released
by the authors.
We generate 2{,}000 five-hop entailment instances with the default
parameters.

Each generated instance includes the same natural-language premises
(describing a fictional ontology) and query, paired with
(i) a natural-language chain-of-thought reasoning trace and
(ii) a sequence of first-order logic proof steps produced by the
generator.
We treat these as the natural-language and symbolic proof views,
respectively.

We randomly split the 2{,}000 instances into 1{,}000 training,
500 development, and 500 test examples.
The 1{,}000 training instances are used to estimate the multi-view
logical subspaces, the 500 development instances are used for
hyperparameter tuning (e.g., steering layer and $\lambda$), and the
500 test instances are used solely for evaluation.
\paragraph{Example.}
Below we show a toy PrOntoQA-style instance with its two views:
a natural-language proof-style explanation and a corresponding
symbolic logic proof trace.

\medskip
\noindent\textbf{Natural-language (NL) proof view}
\begin{verbatim}
Premises:
All flurps are red.
All red things are warm.
Mip is a flurp.

True or false: Mip is warm.

Mip is a flurp.
All flurps are red.
So Mip is red.
All red things are warm.
So Mip is warm.

The query is True.
\end{verbatim}

\medskip
\noindent\textbf{Symbolic proof view}
\begin{verbatim}
Premises:
All flurps are red.
All red things are warm.
Mip is a flurp.

True or false: Mip is warm.

flurp(Mip)
![X1]:((flurp(X1) => red(X1)))
red(Mip)
![X1]:((red(X1) => warm(X1)))
warm(Mip)

The query is True.
\end{verbatim}
\subsection{ProofWriter (OWA and CWA)}

ProofWriter~\citep{proofwriter} provides multi-hop entailment questions
under both open-world (OWA) and closed-world (CWA) semantics, together
with gold natural-language proofs.\footnote{Under OWA, facts not stated
in the context may be unknown; under CWA, unstated facts are treated as
false.}

In this work we use ProofWriter in two ways.
First, for learning multi-view subspaces we focus on the 5-hop regime.
For the OWA setting, we sample 2{,}000 5-hop training
instances from the released data and
a fixed random seed.
Each such instance includes (i) a natural-language proof explaining the
entailment decision.
We translate each gold natural-language proof into a TPTP-style \citep{tptp} symbolic
proof using a GPT-5 few-shot prompt and treat the resulting pair as the
natural-language and symbolic views for subspace estimation.

For subspace learning, we split the 2{,}000 translated 5-hop instances
into 1{,}500 training and 500 development examples.
Training examples are used to estimate the multi-view logical subspaces,
and the remaining 500 5-hop examples are reserved for qualitative and
diagnostic analysis only (see Section~\ref{sec:analysis}); they are never
used for hyperparameter selection or test-time evaluation.

For steering hyperparameters, we operate in the same 3-hop regime as used
at test time. From the official 3-hop validation splits, we subsample
500 OWA and 501 CWA instances with approximately balanced labels and use
them solely to select the steering layer and $\lambda$.

For final evaluation, we focus on the 3-hop subset of ProofWriter, which offers a stable
and interpretable regime for assessing CoT-based reasoning.
From the official 3-hop test splits, we subsample another 500 OWA and
501 CWA instances (again  balanced in labels), and these
held-out 3-hop instances are used exclusively for reporting final
accuracy in our main experiments.
OWA and CWA are reported as separate evaluation settings in all result
tables.
\paragraph{Example (ProofWriter: NL vs.\ symbolic view).}
Below we show a toy ProofWriter-style instance with its two views:
a natural-language proof-style explanation and a corresponding
first-order logic proof trace.

\medskip
\noindent\textbf{Natural-language (NL) proof view}
\begin{quote}\small\ttfamily
Facts:
Alice is young.
All young people are students.
All students are diligent.

Query: Alice is diligent.

Proof:
Alice is young.
All young people are students.
Therefore Alice is a student.
Alice is a student. 
All students are diligent.
Therefore Alice is diligent.

The query is True.
\end{quote}

\medskip
\noindent\textbf{Symbolic proof view}
\begin{quote}\small\ttfamily
Facts:
young(Alice).
![X]:((young(X) => student(X))).
![X]:((student(X) => diligent(X))).

Query: diligent(Alice).

Proof:
young(Alice).
![X]:((young(X) => student(X))).
student(Alice).
student(Alice). 
![X]:((student(X) => diligent(X))).
diligent(Alice).

The query is True.
\end{quote}

\subsection{Few-Shot Examples}

  We use 3-shot in-context learning with one example for each label (True, False, Uncertain). Below we show one representative example.

  \begin{tcolorbox}[breakable, title=Few-Shot Example, colback=white]
  \textbf{Premises:}\\
  All eels are fish.\\
  No fish are plants.\\
  Everything displayed in the collection is either a plant or an animal.\\
  All multicellular animals are not bacteria.\\
  All animals displayed in the collection are multicellular.\\
  A sea eel is displayed in the collection.\\
  The sea eel is an eel or an animal or not a plant.

  \textbf{Hypothesis:}\\
  The sea eel is bacteria.

  \textbf{Reasoning:}\\
  Let's analyze step by step:\\
  1. From premise 6: The sea eel is displayed in the collection.\\
  2. From premise 3: Since it's displayed, the sea eel is either a plant or an animal.\\
  3. From premise 7: The sea eel is an eel or an animal or not a plant.\\
  4. From premise 1: All eels are fish.\\
  5. From premise 2: No fish are plants. So if the sea eel is an eel, it's a fish, thus not a plant.\\
  6. From premise 3: Since it's displayed and not a plant, the sea eel must be an animal.\\
  7. From premise 5: All animals displayed in the collection are multicellular.\\
  8. So the sea eel is a multicellular animal.\\
  9. From premise 4: All multicellular animals are not bacteria.\\
  10. Therefore, the sea eel is NOT bacteria.

  \textbf{Truth value: False}
  \end{tcolorbox}
\section{Generalization Setup}
\label{app:gen-setup}

This appendix provides additional details for the cross-dataset
generalization experiments from PrOntoQA to LogiQA~2.0 reported in
Section~\ref{sec:generalization}. We use an analogous transfer protocol for
{ReClor}, reusing the same fixed PrOntoQA-derived subspace without
retraining.

\subsection{Source of the Steering Subspace}

For all LogiQA~2.0 results, the multi-view logical subspaces are
estimated \emph{exclusively} from synthetic PrOntoQA data.
Specifically, we use the official PrOntoQA code to generate
5{,}000 fictional 5-hop entailment instances in the
\textsc{Fictional-Composed-Rule} setting and run the PCA+CCA pipeline
from Section~\ref{sec:method-subspace} on the pooled residual
activations of paired natural-language and symbolic proofs.
For each model and each layer~$\ell$, this yields an orthonormal
NL-side basis $U^{(\ell)}$ and the corresponding projector
$P^{(\ell)} = U^{(\ell)} {U^{(\ell)}}^\top$.

No LogiQA~2.0 examples are used when fitting PCA, running CCA, or
selecting the subspace dimension~$k$.
As in our in-domain experiments, we restrict steering to a single
layer~$\ell^\star$ per model, chosen based on the mean canonical
correlation criterion described in
Appendix~\ref{app:exp-steer-layer}.
The choice of $\ell^\star$ is fixed before any LogiQA~2.0 evaluation.

\subsection{LogiQA~2.0 Tasks and Splits}

LogiQA~2.0 is a logical reading comprehension benchmark derived from
Chinese civil service examinations and provides two evaluation
settings that we consider:

\begin{itemize}
    \item \textbf{MCR (multiple-choice reading)}:
    given a passage, a question, and four options (A--D), the model
    must select the correct answer;
    \item \textbf{NLI (natural language inference)}:
    given a set of premises and a conclusion, the model must decide
    whether the conclusion is logically supported.
\end{itemize}

For each model and each setting (MCR/NLI), we tune the steering
strength~$\lambda$ on the official LogiQA~2.0 development split and
evaluate on a subset of the official test split.
Concretely, we randomly sample 500 test examples for MCR and another
500 test examples for NLI from the corresponding LogiQA~2.0 test sets
and report accuracy on these held-out subsets.
LogiQA~2.0 examples are therefore used only to select $\lambda$ and
for evaluation; they never participate in subspace estimation.

\subsection{Prompting and Decoding on LogiQA~2.0}

We use the same general Chain-of-Thought prompting protocol as in our
main experiments (Section~\ref{sec:experiments}), adapted to the two
LogiQA~2.0 formats.
All experiments use greedy decoding
(\texttt{do\_sample}~=~\texttt{False}) with a maximum of 1024 new
tokens; generations that fail to produce a valid final-line tag within
this budget are counted as incorrect.

For the MCR setting we use the following prompt template:

\begin{tcolorbox}[breakable, title=LogiQA 2.0 MCR prompt, colback=white]
\textbf{System:} You are a careful logician. Use classical deductive reasoning.

\medskip
\textbf{User:}

Passage:\\
\{text\}

Question: \{question\}

Options:\\
A. \{option\_a\}\\
B. \{option\_b\}\\
C. \{option\_c\}\\
D. \{option\_d\}

Instructions:
\begin{itemize}
    \item First, reason step by step.
    \item Then, on the last line, output exactly:\\
    \texttt{Answer: <A|B|C|D>}
\end{itemize}
\end{tcolorbox}

For the NLI setting we use the following template:

\begin{tcolorbox}[breakable, title=LogiQA 2.0 NLI prompt, colback=white]
\textbf{System:} You are a careful logician. Use classical deductive reasoning.

\medskip
\textbf{User:}

Premises:\\
\{premises\}

Conclusion:\\
\{conclusion\}

Instructions:
\begin{itemize}
    \item First, reason step by step about whether the conclusion follows from the premises.
    \item Then, on the last line, output exactly:\\
    \texttt{Answer: <Entailed|Not Entailed>}
\end{itemize}
\end{tcolorbox}

These templates mirror the CoT prompts used for our other benchmarks
(Appendix~\ref{app:prompts}) but are adapted to the passage--question
format and label space of LogiQA~2.0.

\subsection{Steering Protocol and Hyperparameters}

At inference time on LogiQA~2.0, we reuse the projector
$P^{(\ell^\star)}$ learned from PrOntoQA and apply the same
inference-time steering rule as in Eq.~\eqref{eq:steer}.
For each generated CoT token at time step~$t$, we replace the residual
vector $h_t^{(\ell^\star)}$ by
\[
\tilde h^{(\ell^\star)}_{t}
= h^{(\ell^\star)}_{t}
+ \lambda \,
  \frac{P^{(\ell^\star)} h^{(\ell^\star)}_{t}}
       {\big\|P^{(\ell^\star)} h^{(\ell^\star)}_{t}\big\|_2 + \varepsilon}
  \, \big\|h^{(\ell^\star)}_{t}\big\|_2,
\]
with a small $\varepsilon$ added for numerical stability.
We intervene only on tokens in the generated CoT; the encoding of the
passage, question, premises, and conclusion is left unchanged.

For each model and each LogiQA~2.0 setting (MCR/NLI), we perform a
small grid search over
\[
\lambda \in \{0.02, 0.04, 0.06, 0.08, 0.10, 0.12\}
\]
on the development split and select the value that maximizes CoT
accuracy.
The selected~$\lambda$ is then fixed for all test examples.
No PCA or CCA is ever recomputed on LogiQA~2.0; the subspace is
entirely inherited from PrOntoQA.
\section{Models}
\label{app:models}

\begin{table}[H]
\centering
\small
\begin{tabular}{|l|c|c|}
\hline
Model                     & Params & Reference           \\ \hline
Llama 3.1 8B Instruct     & 8B     & \citet{llama3}   \\
Llama 3.2 3B Instruct     & 3B     & \citet{llama3}  \\
Llama 2 13B Chat          & 13B    & \citet{llama2}         \\
Gemma 2 9B IT             & 9B     & \citet{gemma2}        \\
Phi-3-mini-4k-instruct    & 3.8B   & \citet{phi3}           \\ \hline
\end{tabular}
\caption{Open-weight decoder-only models used in our experiments.}
\label{tab:models}
\end{table}
\section{Prompt Templates}
\label{app:prompts}

In this appendix we list the exact prompt templates used in our
experiments. For models that support a \texttt{system} role (e.g.,
LLaMA and Phi-3 families), we send the \emph{System} and \emph{User}
messages below as two separate chat turns. For Gemma, which does not
expose a system role, we concatenate the system content and the user
content into a single user message.

\subsection{FOLIO Evaluation Prompt}

\begin{tcolorbox}[breakable, title=System message]
You are a helpful reasoning assistant.
\end{tcolorbox}

\begin{tcolorbox}[breakable, title=User message]
You are a careful logician. Use classical deductive reasoning.

\textbf{Premises:}\\
\{premises\}

\textbf{Hypothesis:}\\
\{hypothesis\}

\textbf{Instructions:}
\begin{itemize}
    \item First, reason step by step.
    \item Then, on the last line, output exactly:\\
    Truth value:<True|False|Uncertain>
\end{itemize}
\end{tcolorbox}
\subsection{PrOntoQA Evaluation Prompt}
\label{app:prompt-prontoqa}

\begin{tcolorbox}[breakable, title=System Message ]
You are a helpful reasoning assistant.
\end{tcolorbox}

\begin{tcolorbox}[breakable, title=User Message]
You are a careful logician. Use classical deductive reasoning.

  \textbf{Premises:}\\
  \{premises\}

  True or false: \{query\}

\textbf{Instructions:}
\begin{itemize}
    \item First, reason step by step.
    \item Then, on the last line, output exactly:\\
    Truth value: \textless True\textbar False\textgreater
\end{itemize}
\end{tcolorbox}
  \subsection{ProofWriter (CWA) Evaluation Prompt}

  \begin{tcolorbox}[breakable, title=System Message]
  You are a helpful reasoning assistant.
  \end{tcolorbox}

  \begin{tcolorbox}[breakable, title=User Message]
  You are a careful logician. Use classical deductive reasoning.

  \textbf{Facts:}\\
  \{facts\}

  \textbf{Rules:}\\
  \{rules\}

  \textbf{Query:} \{query\}

  \textbf{Instructions:}
  \begin{itemize}
      \item First, reason step by step. (MAXIMUM 25 steps)\footnote{We set a 25-step limit to prevent runaway generation;
  this is well above the maximum proof depth in ProofWriter.}
      \item Then, on the last line, output exactly:\\
      Truth value: <True|False>
      \item IMPORTANT: If you cannot complete reasoning in 25 steps,
      you MUST output your best judgment anyway on the last line.
  \end{itemize}
  \end{tcolorbox}
  \subsection{ProofWriter (OWA) Evaluation Prompt}

  \begin{tcolorbox}[breakable, title=System Message]
  You are a helpful reasoning assistant.
  \end{tcolorbox}

  \begin{tcolorbox}[breakable, title=User Message]
  You are a careful logician. Use classical deductive reasoning.

  \textbf{Facts:}\\
  \{facts\}

  \textbf{Rules:}\\
  \{rules\}

  \textbf{Query:} \{query\}

  \textbf{Instructions:}
  \begin{itemize}
      \item First, reason step by step (MAXIMUM 25 steps).
      \item Then, on the last line, output exactly:\\
      Truth value: <True|False|Uncertain>
      \item IMPORTANT: If you cannot complete reasoning in 25 steps,
      you MUST output your best judgment anyway on the last line.
  \end{itemize}
  \end{tcolorbox}
\section{Additional Analysis Setup}
\label{app:analysis-setup}

\paragraph{Layer-wise NL--symbolic alignment.}
For Section~\ref{sec:layer-dynamics}, we estimate layer-wise PCA+CCA
mappings between pooled NL and symbolic proof activations on the
PrOntoQA \textsc{Fictional-5hop} training split, and compute mean
canonical correlations $\bar{\rho}^{(\ell)}$ on 500 held-out validation
instances using proof tokens only.

\paragraph{Projection energy and correctness.}
For Section~\ref{sec:energy-signal}, we use {Llama-3.1-8B} and
reuse the same layer-wise projectors $\{P^{(\ell)}\}$ learned on the
PrOntoQA training split.
On 500 PrOntoQA test instances we generate zero-shot Greedy-CoT traces
(no steering) and take the mean projection energy of CoT tokens at each
layer as an unsupervised confidence score for ROC-AUC evaluation.

\paragraph{Behavioral comparison of Greedy vs.\ Steered.}
For Section~\ref{sec:behavioral_analysis}, we again use the same 500
PrOntoQA test instances and generate two CoTs per instance:
Greedy-CoT and \ourmethod{} with layer-16 steering and $\lambda=0.08$.
Step counts are defined as the number of lines in the explanation before
the final \texttt{Truth value:} line, and all length and lexical
statistics are computed over these explanation spans only.
The lexicons used in the style analysis are listed in
Appendix~\ref{app:style-lexicons}.
\section{Additional Experimental Details}
\label{app:exp-details}

\subsection{Selecting Steering Layers}
\label{app:exp-steer-layer}

In our main experiments we restrict steering to a single layer
$\ell^\star$ for each model–benchmark pair.
Rather than sweeping over all decoder layers, we select a small set
of candidate layers based on the strength of the learned multi-view
alignment.

Concretely, let $L$ denote the number of decoder layers of a given
model.
Using the training split for a benchmark, we first estimate the
multi-view logical subspace at every layer $\ell \in \{1,\dots,L\}$
(Section~\ref{sec:method-subspace}) and compute a scalar score
$\bar\rho^{(\ell)}$ defined as the mean canonical correlation of the
top-$k$ canonical components.
We then form a candidate set $\mathcal{C}$ by taking the $8$ layers
with the highest $\bar\rho^{(\ell)}$ in the
middle-to-upper part of the network.

For each model–benchmark pair, we perform a small grid search over
$\ell \in \mathcal{C}$ and
\[
\lambda \in \{0.02, 0.04, 0.06, 0.08, 0.10, 0.12, 0.14\}
\]
on a held-out development set, and select the combination
$(\ell^\star, \lambda)$ that maximises CoT accuracy.
Once selected, $(\ell^\star, \lambda)$ is fixed for all test
instances for that model and benchmark.
Empirically, the chosen steering layers for different models and
tasks always fall in the middle-to-upper range of the stack, where
NL–symbolic alignment is strongest.
\subsection{Corpus and tagging for energy analysis}
\label{app:energy-data}

For the token-level energy analyses in Section~\ref{sec:analysis},
we construct a synthetic rule-based corpus using the official
PrOntoQA generation code.

\paragraph{Corpus construction.}
We generate 5{,}000 fictional 5-hop instances using the
\texttt{Fictional-5hop} configuration with composed rule option.

\paragraph{Token categorization.}
We assign each proof token to one of six coarse categories:
negation, quantifier, copula, entity, concept, and structure.
This is done using a small hand-crafted lexicon and simple
pattern-based heuristics; tokens that do not match any category are
ignored in the energy aggregations.

\paragraph{Subspace for analysis.}
For a fixed model and layer, we estimate the multi-view logical
subspace $U^{(\ell)}$ on the full 5{,}000-instance corpus and then
reuse this subspace when computing all token-level and per-category
energies in Section~\ref{sec:analysis}.

\paragraph{Lexicon for token categories.}
For reproducibility, we list the exact lexical sets used to tag tokens
in the synthetic PrOntoQA-style corpus into the six coarse categories
(entities, concepts, properties, quantifiers, negation, and structure).

\smallskip
\noindent\textbf{Entity names.}
We tag the following surface forms as entities:
\texttt{fae}, \texttt{rex}, \texttt{sally}, \texttt{max},
\texttt{alex}, \texttt{sam}, \texttt{polly}, \texttt{stella},
\texttt{wren}.

\smallskip
\noindent\textbf{Concept names.}
We tag the following nonce nouns (including both singular and plural
forms) as concept tokens:
\texttt{wumpus}, \texttt{yumpus}, \texttt{zumpus}, \texttt{dumpus},
\texttt{rompus}, \texttt{numpus}, \texttt{tumpus}, \texttt{vumpus},
\texttt{impus}, \texttt{jompus}, \texttt{gorpus}, \texttt{shumpus},
\texttt{lempus}, \texttt{sterpus}, \texttt{grimpus}, \texttt{lorpus},
\texttt{brimpus},
\texttt{timpus}, \texttt{yimpus}, \texttt{rempus}, \texttt{fompus},
\texttt{worpus}, \texttt{terpus}, \texttt{gerpus}, \texttt{kerpus},
\texttt{scrompus}, \texttt{zhorpus}, \texttt{bompus}, \texttt{jelpus},
\texttt{felpus}, \texttt{chorpus}, \texttt{hilpus}, \texttt{storpus},
\texttt{yerpus}, \texttt{boompus}, \texttt{gwompus}, \texttt{rorpus},
\texttt{quimpus},
\texttt{wumpuses}, \texttt{yumpuses}, \texttt{zumpuses},
\texttt{dumpuses}, \texttt{rompuses}, \texttt{numpuses},
\texttt{tumpuses}, \texttt{vumpuses}, \texttt{impuses},
\texttt{jompuses}, \texttt{gorpuses}, \texttt{shumpuses},
\texttt{lempuses}, \texttt{sterpuses}, \texttt{grimpuses},
\texttt{lorpuses}, \texttt{brimpuses}.

\smallskip
\noindent\textbf{Properties.}
Adjectives and property words are taken from:
\texttt{blue}, \texttt{red}, \texttt{brown}, \texttt{orange},
\texttt{small}, \texttt{large},
\texttt{metallic}, \texttt{wooden}, \texttt{luminous}, \texttt{liquid},
\texttt{transparent}, \texttt{opaque},
\texttt{nervous}, \texttt{happy}, \texttt{feisty}, \texttt{shy},
\texttt{bright}, \texttt{dull},
\texttt{sweet}, \texttt{sour}, \texttt{spicy}, \texttt{bitter},
\texttt{floral}, \texttt{fruity}, \texttt{earthy},
\texttt{hot}, \texttt{cold}, \texttt{temperate},
\texttt{kind}, \texttt{mean}, \texttt{angry}, \texttt{amenable},
\texttt{aggressive},
\texttt{melodic}, \texttt{muffled}, \texttt{discordant}, \texttt{loud},
\texttt{slow}, \texttt{moderate}, \texttt{fast},
\texttt{windy}, \texttt{sunny}, \texttt{overcast}, \texttt{rainy},
\texttt{snowy}.

\smallskip
\noindent\textbf{Quantifiers.}
We tag the following as quantifier tokens:
\texttt{each}, \texttt{every}, \texttt{all}, \texttt{a}, \texttt{an},
\texttt{no}, \texttt{everything}, \texttt{that}.

\smallskip
\noindent\textbf{Negation.}
We treat \texttt{not} as the sole explicit negation marker in this
corpus.

\smallskip
\noindent\textbf{Structural tokens.}
Finally, we group meta-level and connective words into a coarse
``structure'' category:
\texttt{true}, \texttt{false}, \texttt{the}, \texttt{query}, \texttt{or},
\texttt{and}, \texttt{premises}, \texttt{assume}, \texttt{this},
\texttt{contradicts}, \texttt{with}, \texttt{fact}.
\subsection{Additional Energy Plots Across Models and Layers}
\label{app:energy-viz}

\paragraph{Token-level projection energy.}
Figure ~\ref{fig:category-energy} in the main text shows category-wise
projection energy for a representative middle layer.
Here we provide additional plots for other layers and models.

\begin{figure}[H]
  \centering
  % Llama-3.1-8B, layer 25
  \includegraphics[width=0.9\linewidth]{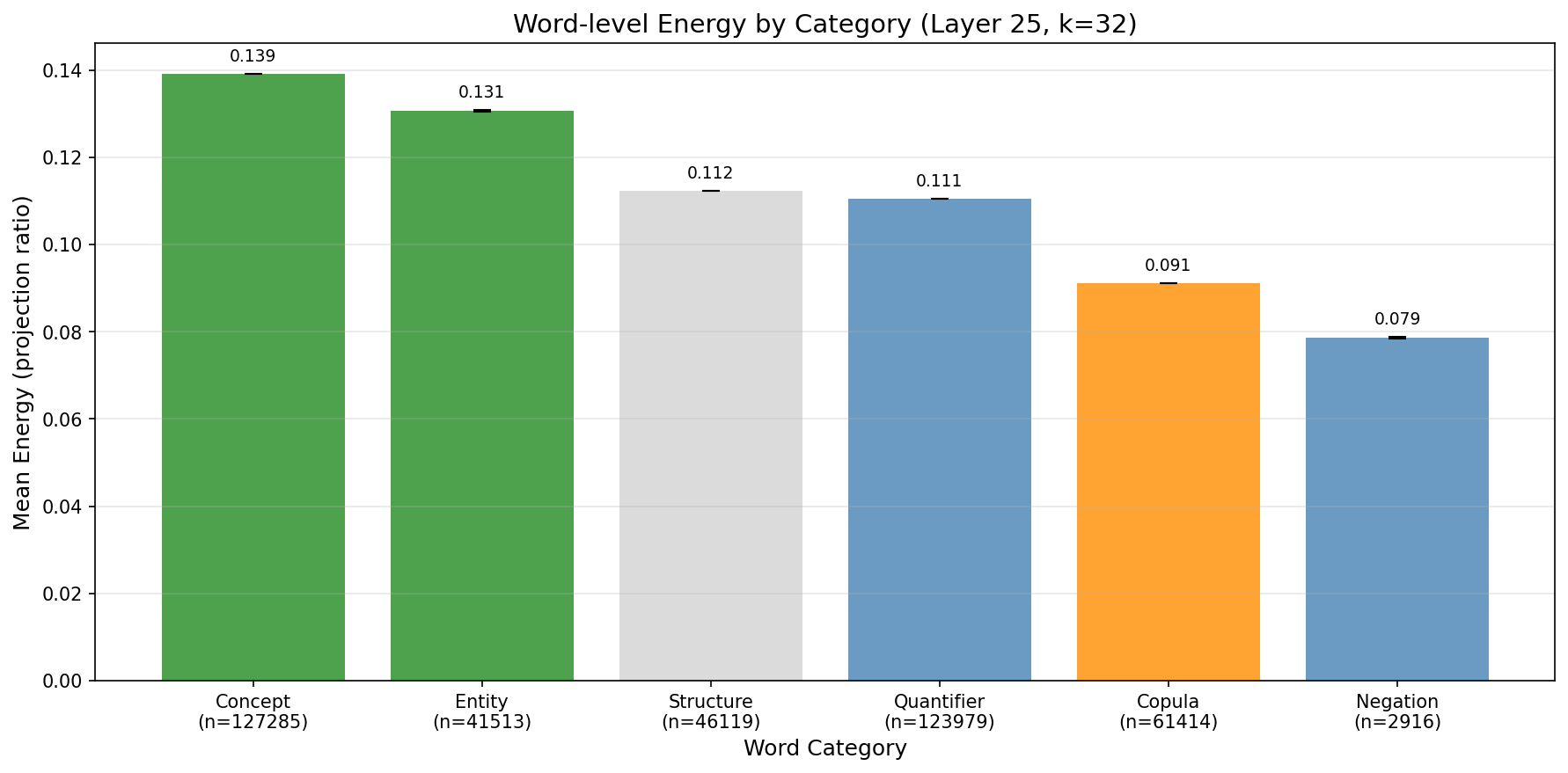}
  \caption{
  Token-level projection energy $E^{(\ell)}(r)$ aggregated by token
  category for {Llama-3.1-8B} at layer 25.
  }
  \label{fig:app-cat-energy-llama25}
\end{figure}

\begin{figure}[H]
  \centering
  % Phi-3-Mini, layers 17 and 18
  \includegraphics[width=0.45\textwidth]{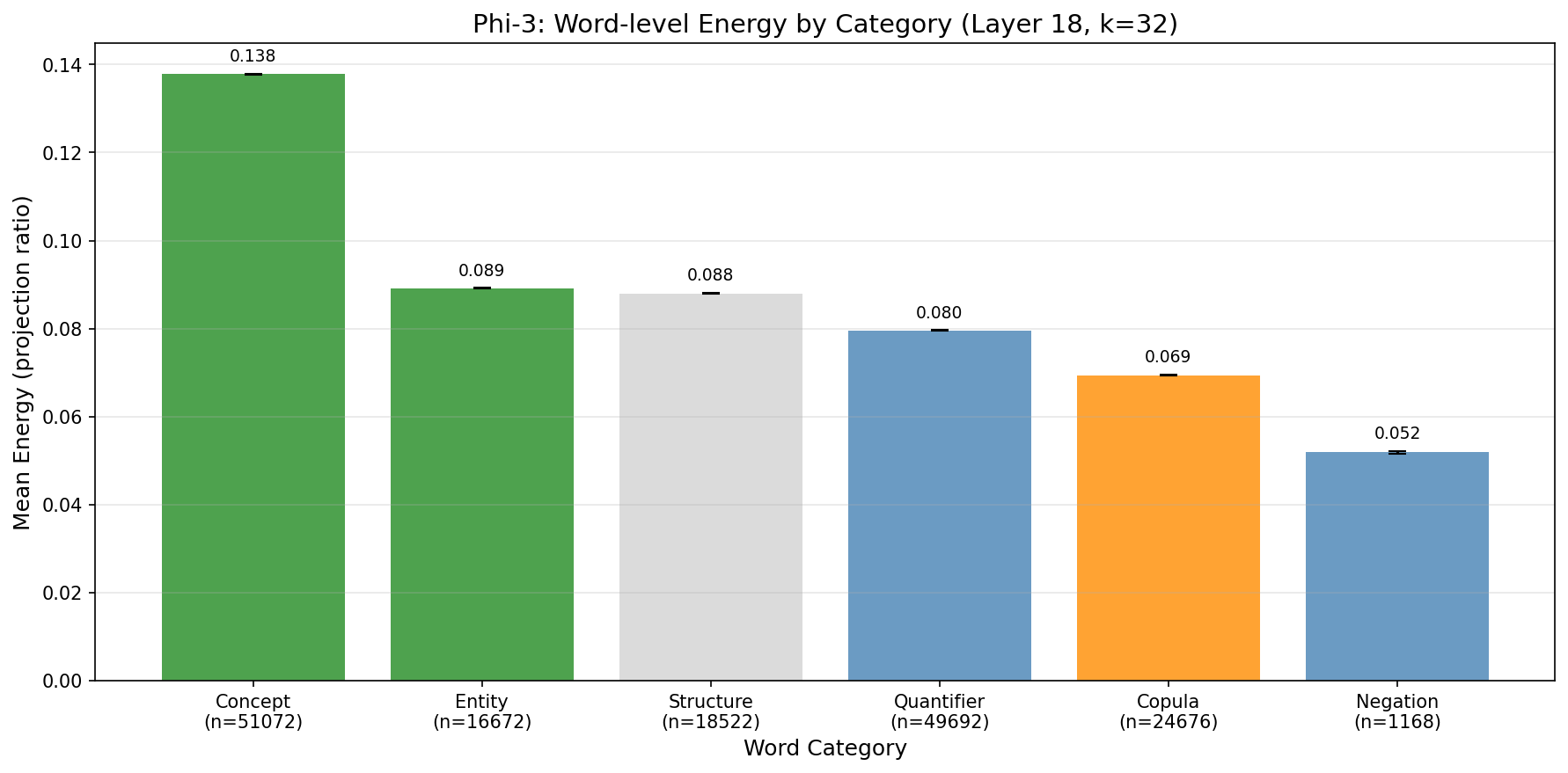}
  
  \caption{
  Token-level projection energy $E^{(\ell)}(r)$ aggregated by token
  category for {Phi-3-Mini} at layers 18.
  }
\end{figure}

\paragraph{Per-direction contribution heatmaps.}
We also visualize the per-direction contributions for additional layers.

\begin{figure}[H]
  \centering
  % Llama-3.1-8B, layer 25
  \includegraphics[width=0.9\linewidth]{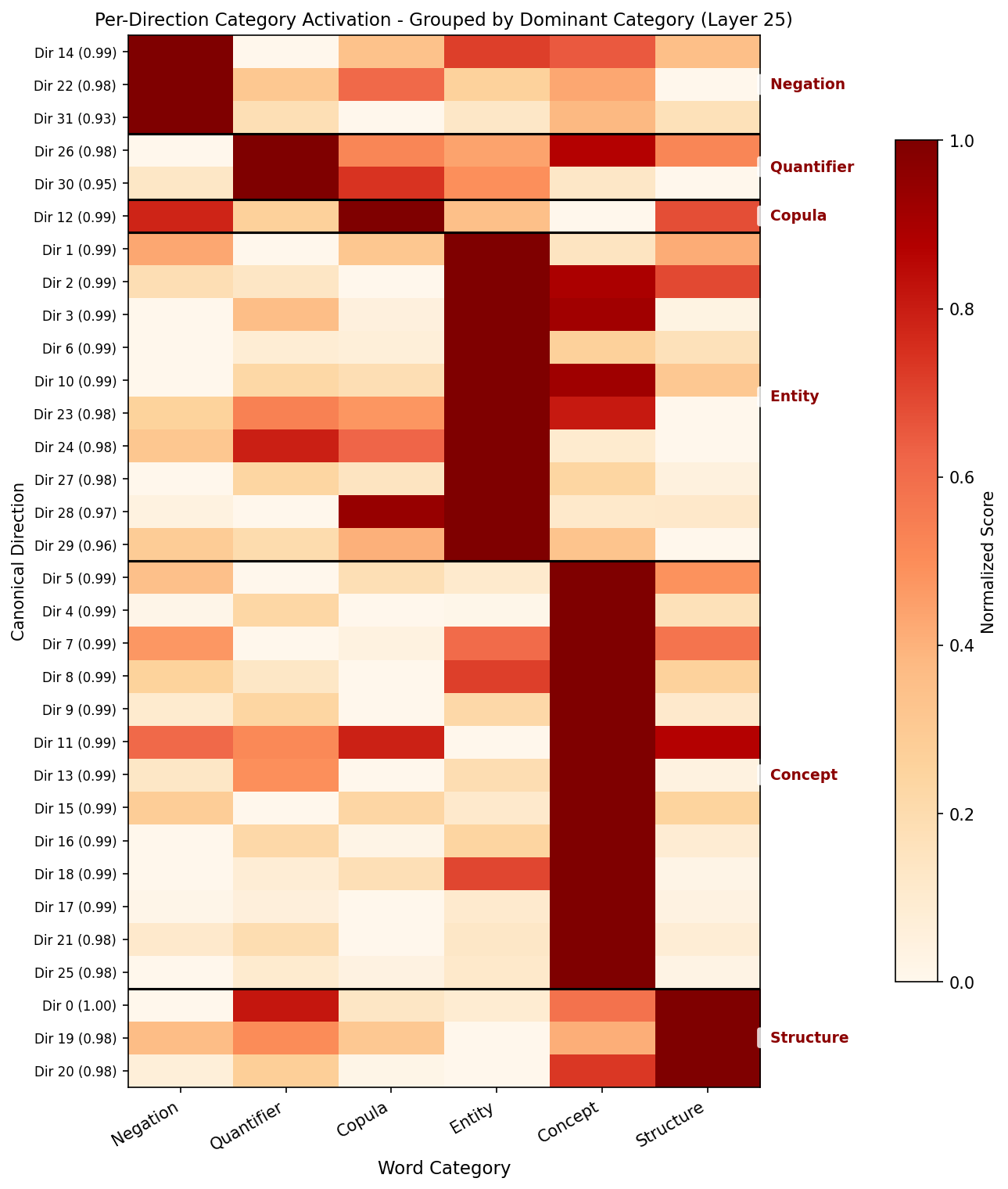}
  \caption{
  Per-direction contribution heatmap for {Llama-3.1-8B} at layer 25.
  Rows correspond to canonical directions and columns to token categories
  (row-normalized).
  }
  \label{fig:app-dir-llama25}
\end{figure}

\begin{figure}[H]
  \centering
  % Phi-3-Mini, layers 17 and 18
  \includegraphics[width=0.45\textwidth]{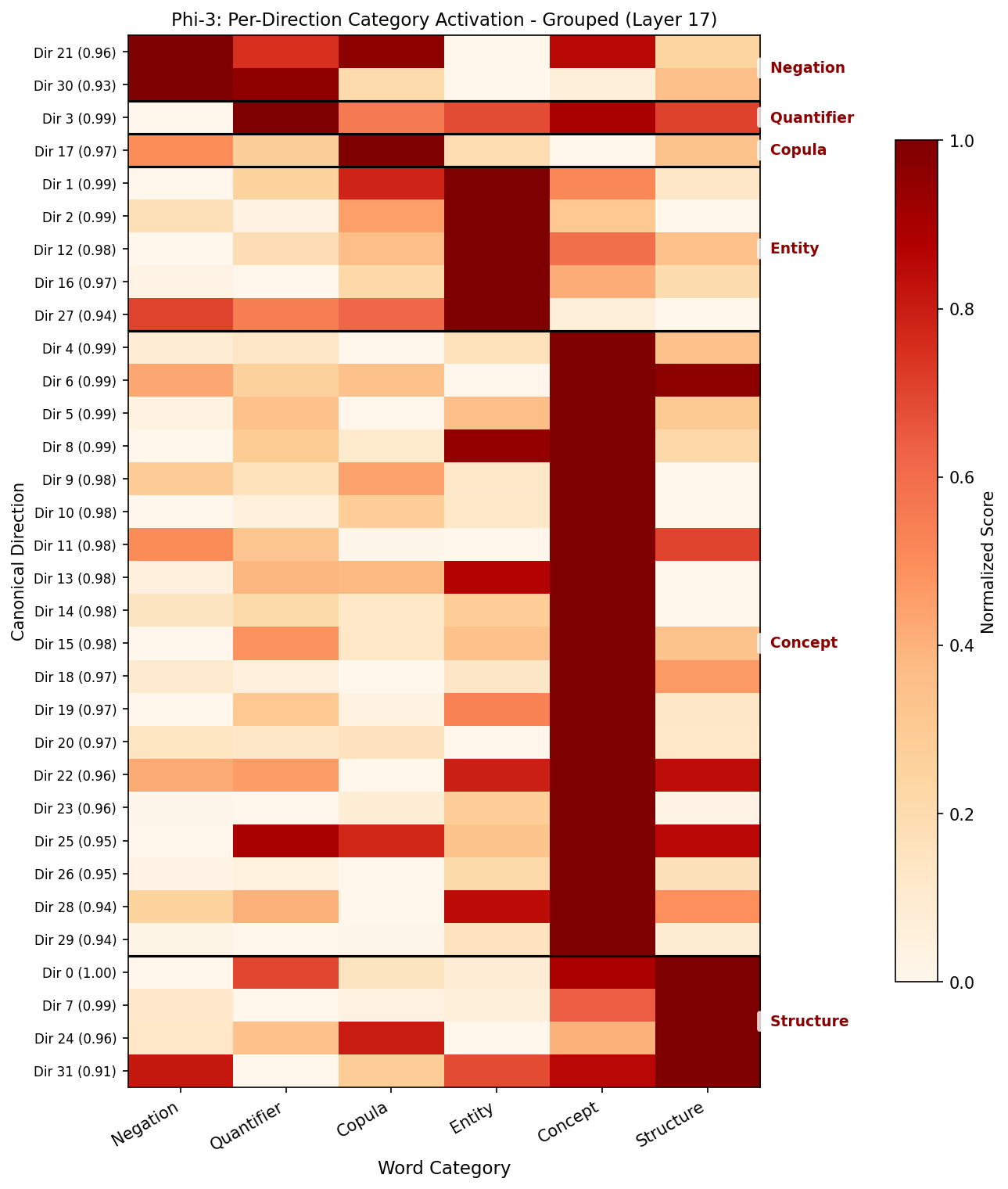}
  
  \caption{
  Per-direction contribution heatmaps for {Phi-3-Mini} at layers
  17 (left) and 18 (right).}
\end{figure}
\subsection{Additional results on a reasoning model}
\label{app:qwen3}

To test whether LSS remains effective on a reasoning-specialized model,
we evaluate \ourmethod{} on {Qwen3-4B} on {PrOntoQA}.
Without steering, {Qwen3-4B} achieves 87.2\% accuracy. With LSS,
the best setting reaches 93.2\% (+6.0 points), showing that steering
still yields substantial gains even when the base model already performs
strongly.

\begin{table}[t]
\centering
\small
\setlength{\tabcolsep}{6pt}
\begin{tabular}{@{}lc@{}}
\toprule
\textbf{Setting ({Qwen3-4B}, PrOntoQA)} & \textbf{Accuracy (\%)} \\
\midrule
No steer & 87.2 \\
LSS ($\lambda=0.02$, layer=20) & 89.8 \\
LSS ($\lambda=0.04$, layer=20) & 89.8 \\
LSS ($\lambda=0.06$, layer=20) & 91.2 \\
LSS ($\lambda=0.08$, layer=20) & 92.0 \\
LSS ($\lambda=0.10$, layer=20) & \textbf{93.2} \\
LSS ($\lambda=0.12$, layer=20) & 93.0 \\
\bottomrule
\end{tabular}
\caption{Additional results on {Qwen3-4B} on {PrOntoQA}. LSS
remains effective on a reasoning-specialized model, with the best setting
improving accuracy from 87.2 to 93.2.}
\label{tab:qwen3}
\end{table}
\section{Additional Layer-wise Analysis}
\label{app:layer-curves}

\begin{figure}[h]
    \centering
   \includegraphics[width=0.48\linewidth]{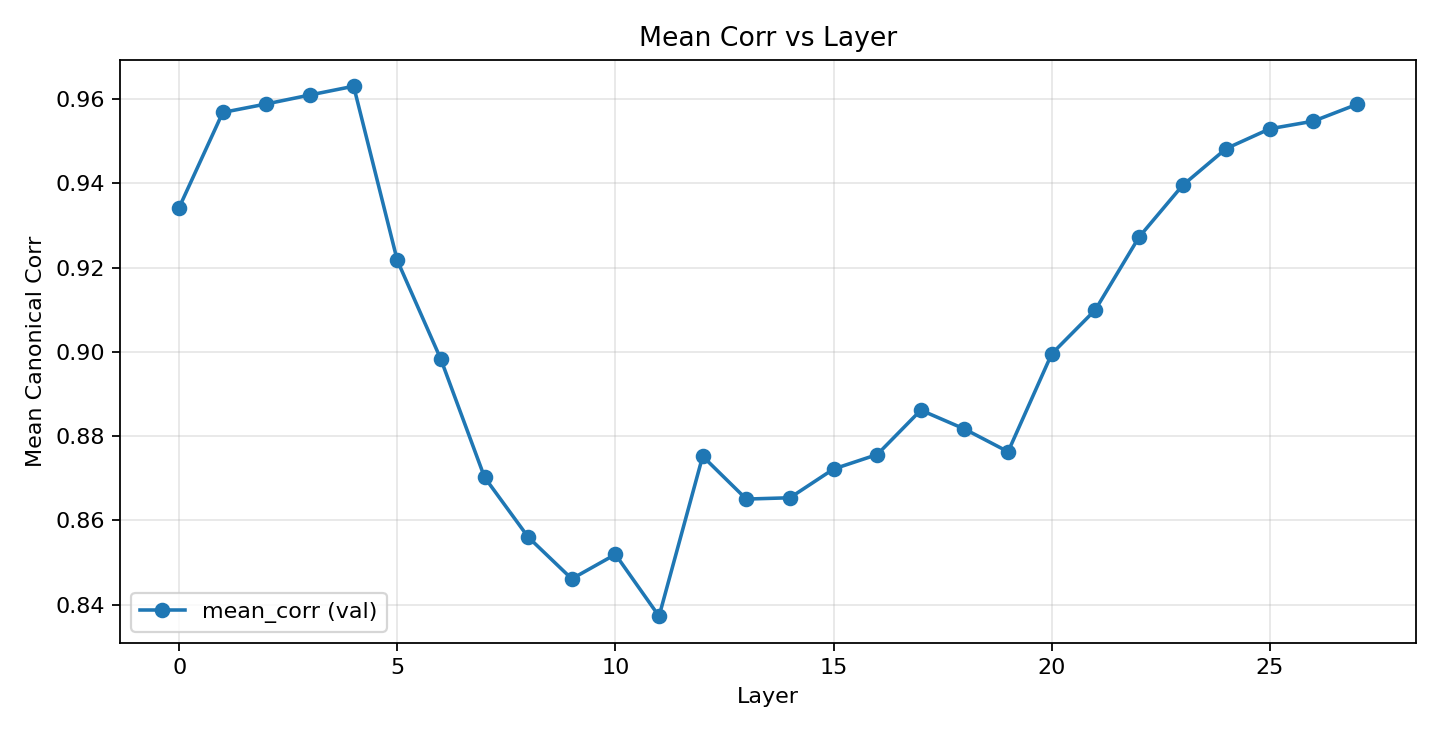}
   \includegraphics[width=0.48\linewidth]{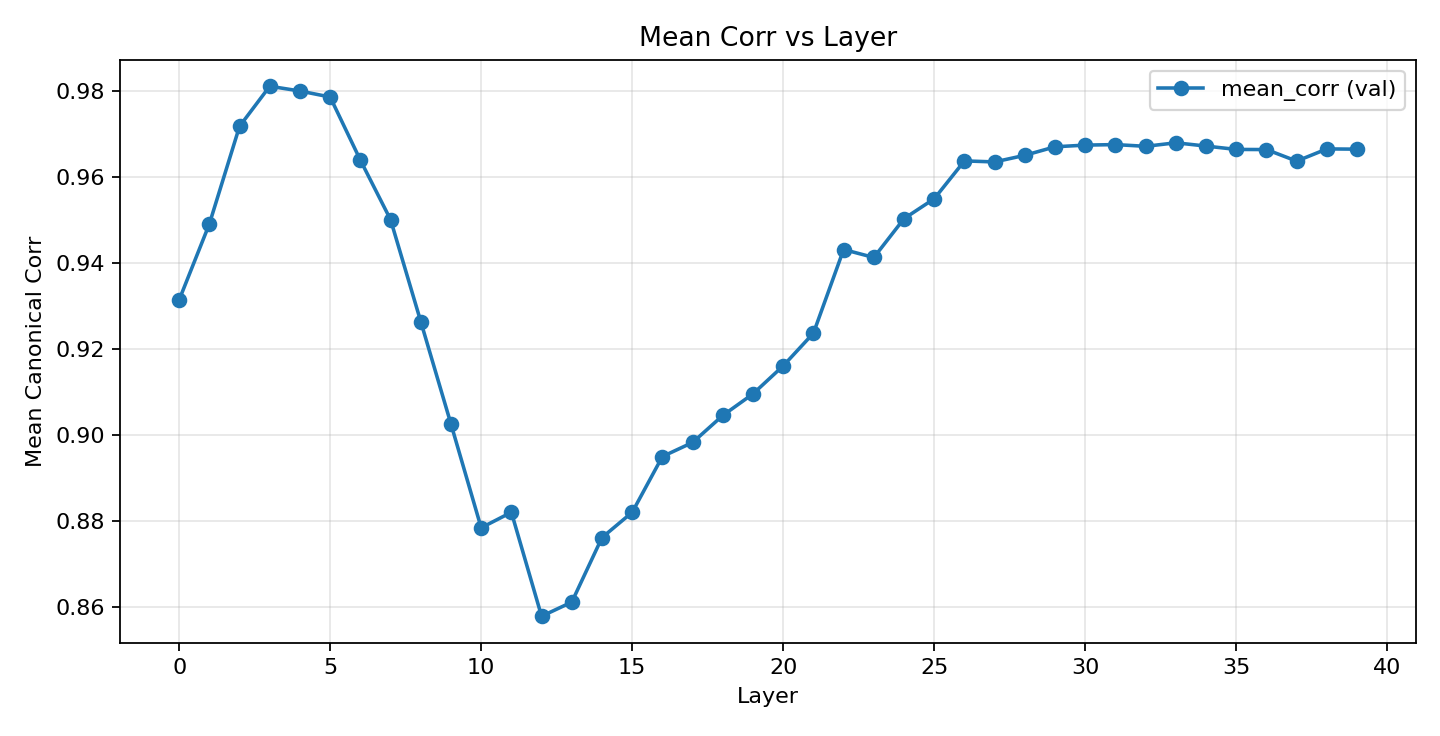}
   \includegraphics[width=0.48\linewidth]{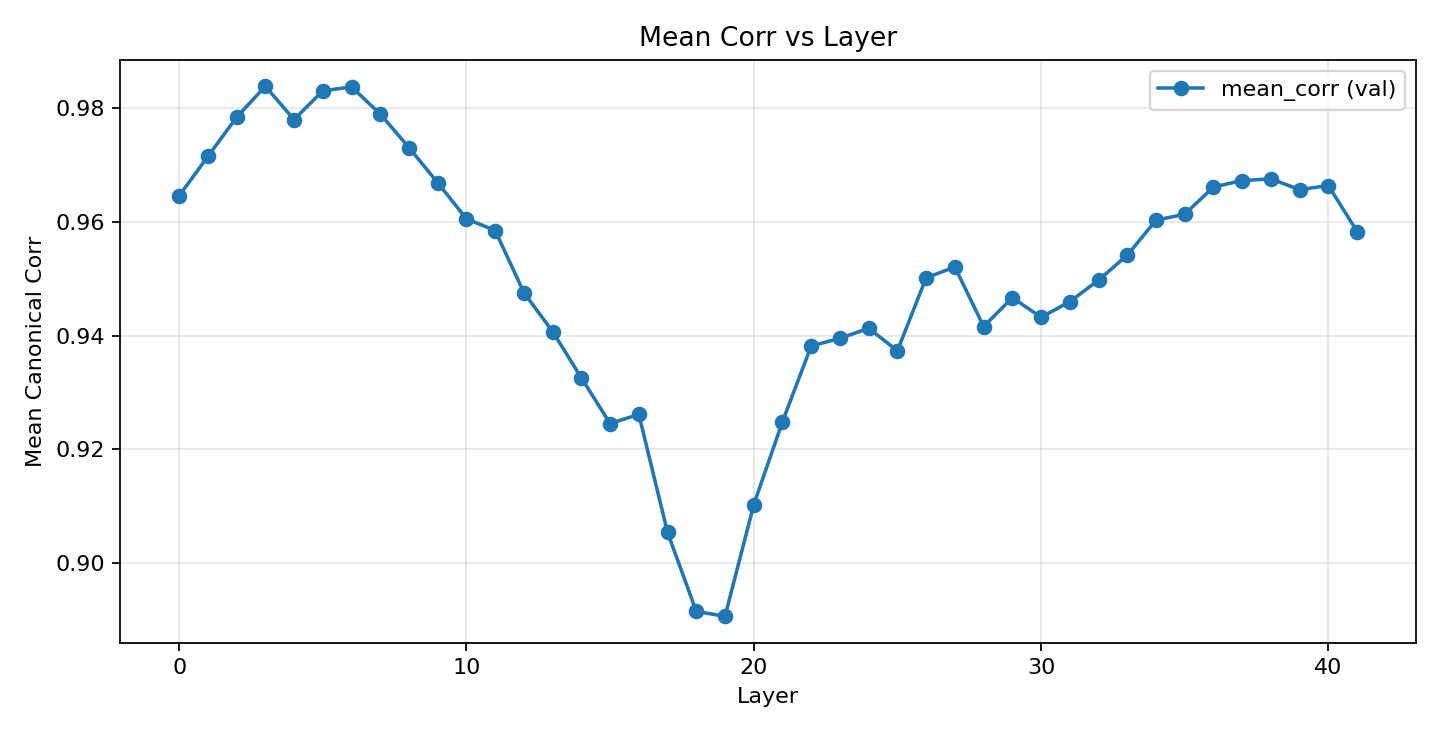}
   \includegraphics[width=0.48\linewidth]{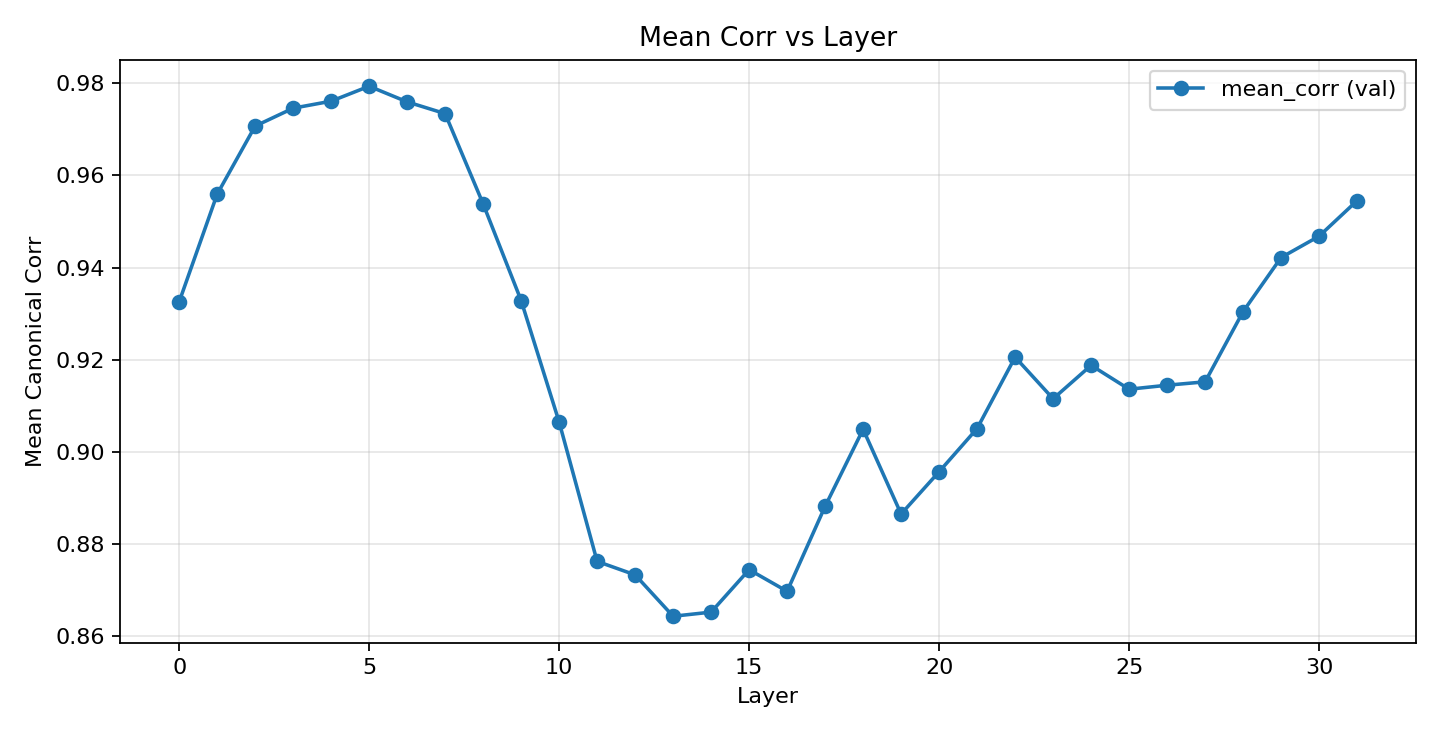}
    \caption{Layer-wise canonical correlation for all evaluated models. The ``high-low-high'' trend is universal.}
    \label{fig:all-layer-curves}
\end{figure}

In Section~\ref{sec:layer-dynamics}, we presented the alignment dynamics for {Llama-3.1-8B}. Figure~\ref{fig:all-layer-curves} presents the corresponding analysis for the remaining models. 
Despite differences in architecture and depth, all models exhibit the characteristic U-shaped trajectory: high initial alignment, a drop in the middle reasoning layers, and a recovery in the final layers.

\section{Details on Correctness Prediction}
\label{app:roc-details}

\begin{figure}[h]
    \centering
    % ROC 曲线图
    \includegraphics[width=0.6\linewidth]{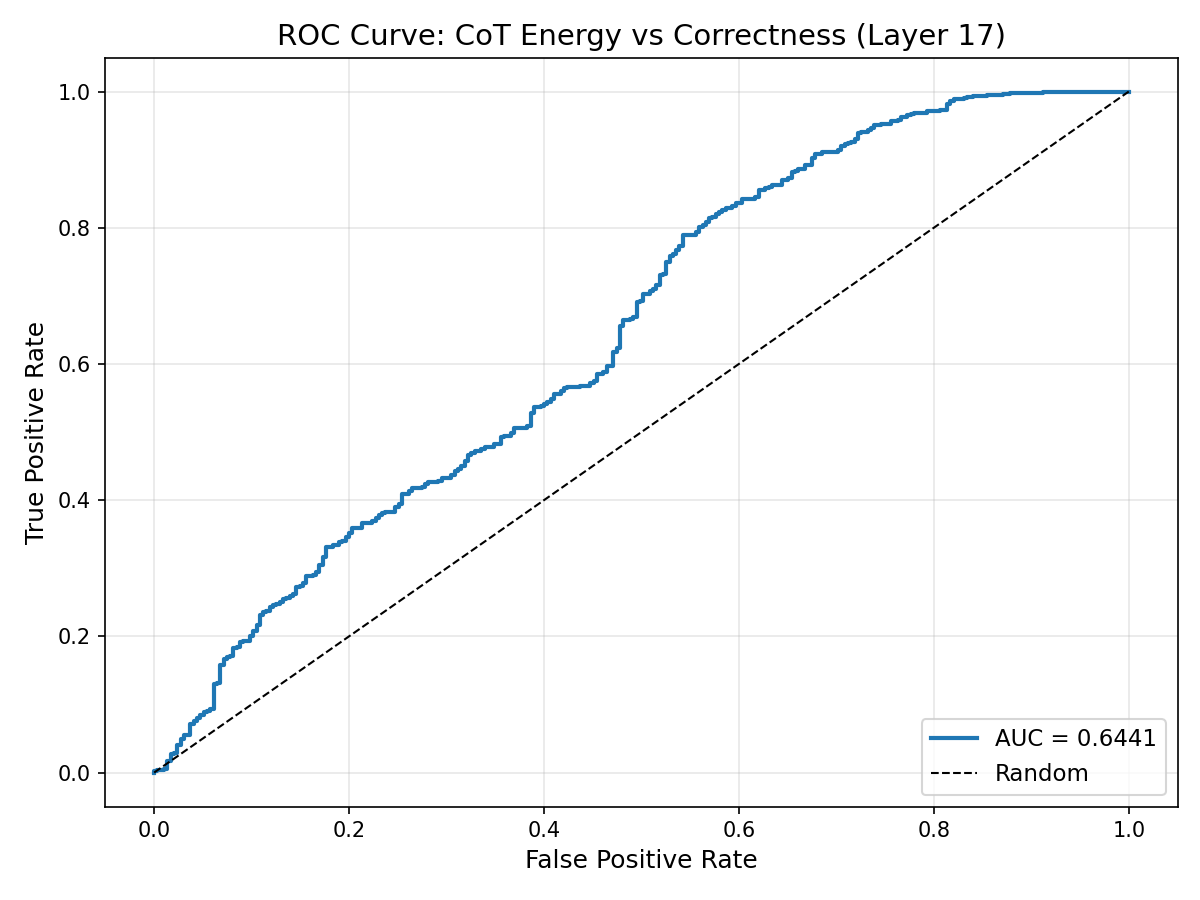}
    \caption{
    \textbf{ROC Curve for Correctness Prediction (Layer 17).} 
    Discriminative performance of the unsupervised projection energy metric (solid line) compared to random chance (dashed line).
    }
    \label{fig:roc-curve}
\end{figure}

In Section~\ref{sec:energy-signal}, we demonstrated that the projection energy in the logical subspace acts as a proxy for reasoning correctness. 
Figure~\ref{fig:roc-curve} provides the detailed Receiver Operating Characteristic (ROC) curve for the best-performing layer (Layer 17) of {Llama-3.1-8B}.

The Area Under the Curve (AUC) is \textbf{0.6441}. While not a perfect classifier, this result is significant given that the metric is entirely \textbf{unsupervised}—derived solely from the geometric properties of the activation space without any training on labeled correctness data.

\section{Lexicons for Style Analysis}
\label{app:style-lexicons}

For the linguistic style analysis in
Section~\ref{sec:behavioral_analysis}, we group tokens into two small
hand-crafted lexicons and report frequency changes at the group level.

\paragraph{Reasoning verbs.}
We treat the following tokens as ``reasoning verbs'':
\texttt{\{know, given, conclude, think, assume\}}.
These mainly occur in metacognitive phrases such as
\emph{``we know that ...''}, \emph{``given that ...''},
or \emph{``we can conclude that ...''}
and typically mark subjective or self-referential reasoning.

\paragraph{Logical connectives.}
We treat the following tokens as ``logical connectives'':
\texttt{\{since, if, then, so, therefore, because\}}.
These terms usually appear in explicit deductive patterns such as
\emph{``since A, ...''}, \emph{``if A then B''},
or \emph{``..., so B''} and directly signal causal structure.

All counts computed as raw token
frequencies over the CoT spans, and percentage changes are reported
relative to the Greedy-CoT baseline.
\begin{table}[t]
\centering
\small
\setlength{\tabcolsep}{3pt} % 缩小列间距
\begin{tabular}{@{}lrrrr@{}}
\toprule
Category / token & Baseline & Steered & $\Delta$ & $\Delta$\% \\
\midrule
\multicolumn{5}{@{}l}{\textbf{Logical connectives}} \\
All connectives      & 4263 & 4466 & +203 & +4.8\% \\
since                & 1800 & 1926 & +126 & +7.0\% \\
if                   & 1128 & 1179 &  +51 & +4.5\% \\
then                 &  641 &  664 &  +23 & +3.6\% \\
so                   &  126 &  149 &  +23 & +18.3\% \\
therefore            &  443 &  447 &   +4 & +0.9\% \\
because              &  123 &  101 &  -22 & $-17.9$\% \\
\midrule
\multicolumn{5}{@{}l}{\textbf{Reasoning verbs}} \\
All reasoning verbs  & 7379 & 6145 & -1234 & $-16.7$\% \\
know                 & 1419 &  944 &  -475 & $-33.5$\% \\
given                & 5176 & 4483 &  -693 & $-13.4$\% \\
conclude             &  770 &  700 &   -70 &  $-9.1$\% \\
\bottomrule
\end{tabular}
\caption{Llama-3.1-8B: lexical changes in CoTs after symbolic-subspace steering. Logical connectives slightly increase overall, while reasoning verbs decrease.}
\label{tab:lexical-story}
\end{table}

\section{Steering Robustness and Directionality}
\label{sec:appendix_steering_robustness}

To verify the robustness and semantic specificity of our steering method, we conducted a detailed sensitivity analysis on \textbf{Llama-3-8B} using the \textbf{ProntoQA} dataset. 
We examined the impact of steering strength ($\lambda$) across different layers (e.g., Layers 16 and 25), explicitly comparing \textbf{our steering method} against a \textbf{randomly generated orthogonal basis}.

\begin{figure}[h]
    \centering
    % 子图 A
    \begin{subfigure}[b]{0.48\linewidth}
        \centering
        \includegraphics[width=\linewidth]{figures/llama_16.png}
        \caption{Layer 16}
        \label{fig:layer16}
    \end{subfigure}
    \hfill
    % 子图 B
    \begin{subfigure}[b]{0.48\linewidth}
        \centering
        \includegraphics[width=\linewidth]{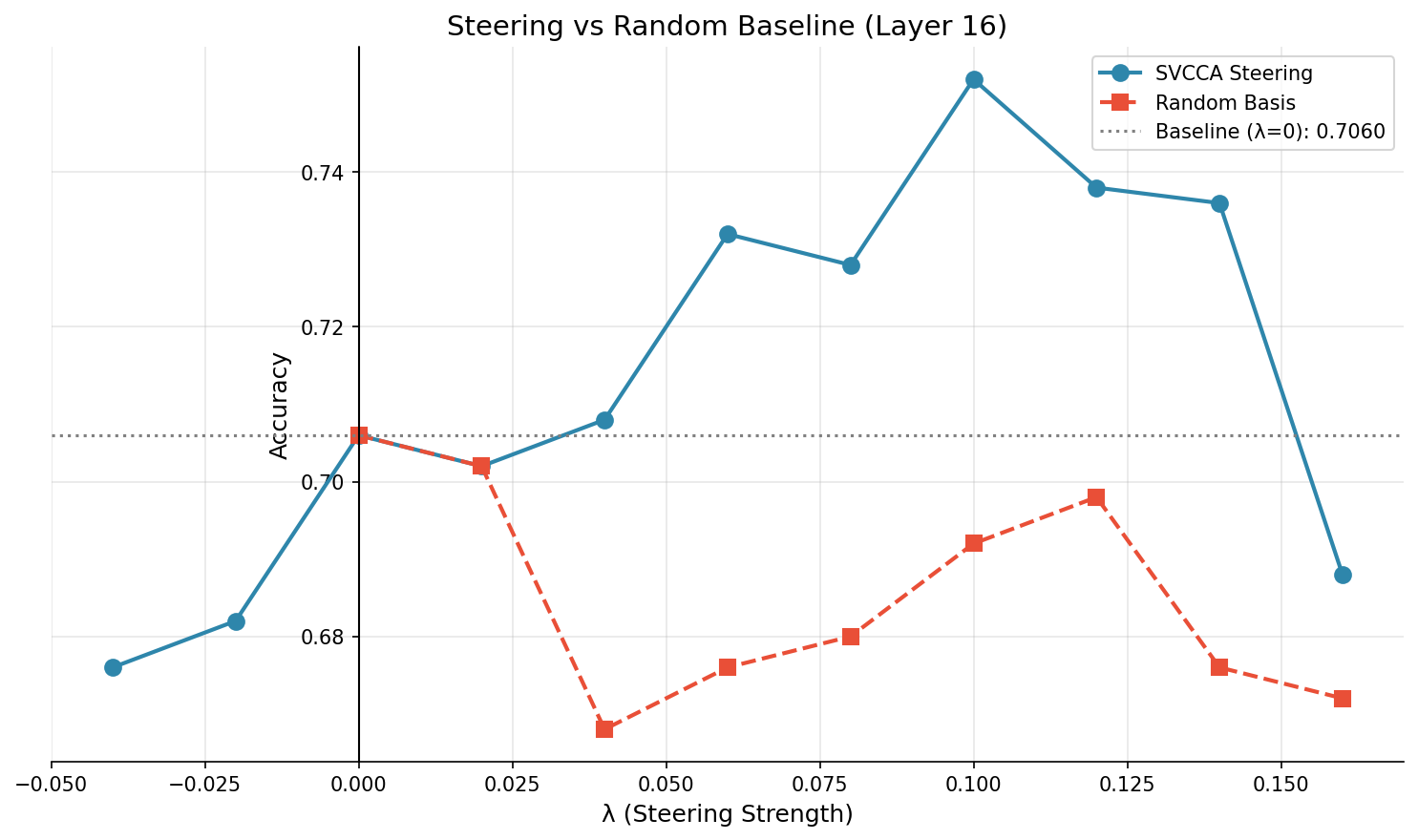}
        \caption{Layer 25}
        \label{fig:layer25}
    \end{subfigure}
    
    \caption{\textbf{Analysis of steering directionality and strength on Llama-3-8B.} We compare Layer 16 (a) and Layer 25 (b). The blue line (\textbf{Our Steering}) shows clear semantic directionality compared to the random baseline (orange).}
    \label{fig:steering_robustness}
\end{figure}

The results (representative plot for Layer 16 shown in Figure~\ref{fig:steering_robustness}) highlight two key findings:
\begin{itemize}
    \item \textbf{Directional Semantics:} The sharp performance drop for negative $\lambda$ values confirms that the extracted vector aligns with a specific, task-relevant semantic direction.
    \item \textbf{Superiority over Random Basis:} Our method consistently outperforms the \textbf{random orthogonal basis}, which exhibits high variance and no meaningful directionality, further validating that the performance gains are driven by the learned subspace structure.
\end{itemize}
\subsection{Robustness to the subspace dimension $k$}
\label{app:k-robustness}

We also study robustness to the number of canonical components used to define
the shared logical subspace. In our main experiments we set $k=32$. To test
whether performance is sensitive to this choice, we train additional shared
subspaces with $k=16$ and $k=64$ on {Llama-3.1-8B} on
{PrOntoQA}, while reusing the same steering layer and $\lambda$ as in
the main experiment (i.e., no retuning).

Table~\ref{tab:k-robustness} shows that performance is not fragile around the
choice of $k$: a smaller subspace ($k=16$) remains competitive, while a much
larger subspace ($k=64$) performs worse, likely because it introduces noisier
or less stable directions. These results support the use of a moderate
low-dimensional shared subspace.

\begin{table}[h]
\centering
\small
\setlength{\tabcolsep}{10pt}
\begin{tabular}{@{}lc@{}}
\toprule
\textbf{$k$} & \textbf{Accuracy (\%)} \\
\midrule
16 & 73.0 \\
32 (main) & \textbf{75.4} \\
64 & 71.0 \\
\bottomrule
\end{tabular}
\caption{Robustness to the subspace dimension $k$ on {PrOntoQA} with
{Llama-3.1-8B}. Additional subspaces with $k=16$ and $k=64$ are trained,
while the steering layer and $\lambda$ are reused from the main experiment
without retuning.}
\label{tab:k-robustness}
\end{table}
\section{Projection Energy Definitions}
\label{app:energy-defs}

\paragraph{Token-level subspace energy.}
For a layer-$\ell$ residual vector $r \in \mathbb{R}^D$ and the
multi-view logical subspace basis
$U^{(\ell)} = [u^{(\ell)}_1,\dots,u^{(\ell)}_k]
\in \mathbb{R}^{D \times k}$,
we define the normalized projection energy
\begin{equation}
E^{(\ell)}(r)
= \frac{\big\| r^\top U^{(\ell)} \big\|_2^2}{\big\|r\big\|_2^2}
= \sum_{j=1}^k \frac{\big(r^\top u^{(\ell)}_j\big)^2}{\big\|r\big\|_2^2}.
\end{equation}
This is the quantity used in Eq.~\eqref{eq:token-energy} in the main
text.

\paragraph{Global per-direction scores.}
For a collection of tokens $\{r_i\}_{i=1}^M$ (e.g., all tokens in a
dataset or all tokens of a given type), we define the global
per-direction score
\begin{equation}
s^{(\ell)}_{\mathrm{global}}(i,j)
= \frac{\big(r_i^\top u^{(\ell)}_j\big)^2}{\big\|r_i\big\|_2^2},
\end{equation}
so that $E^{(\ell)}(r_i) = \sum_{j=1}^k s^{(\ell)}_{\mathrm{global}}(i,j)$.
Averaging $s^{(\ell)}_{\mathrm{global}}(i,j)$ over tokens of a given
class gives an empirical estimate of the fraction of total variance of
that class explained by direction $u^{(\ell)}_j$ (see below).

\paragraph{Subspace-normalized per-direction scores.}
We also consider a version normalized within the subspace:
\begin{equation}
s^{(\ell)}_{\mathrm{subspace}}(i,j)
= \frac{\big(r_i^\top u^{(\ell)}_j\big)^2}
       {\sum_{m=1}^k \big(r_i^\top u^{(\ell)}_m\big)^2},
\end{equation}
which decomposes the token’s subspace energy across the $k$ directions
and satisfies $\sum_{j=1}^k s^{(\ell)}_{\mathrm{subspace}}(i,j) = 1$
whenever $E^{(\ell)}(r_i) > 0$.

\paragraph{Explained-variance perspective.}
At the population level, let $R$ be a random residual vector at layer
$\ell$ with covariance $\Sigma^{(\ell)}$.
The variance of $R$ along direction $u^{(\ell)}_j$ is
$\mathbb{E}[(R^\top u^{(\ell)}_j)^2]$, and the total variance is
$\mathbb{E}[\|R\|_2^2] = \mathrm{trace}(\Sigma^{(\ell)})$.
Thus, averaging $s^{(\ell)}_{\mathrm{global}}(i,j)$ over a set of
tokens of a given type yields an empirical estimate of
\[
\frac{\mathbb{E}[(R^\top u^{(\ell)}_j)^2]}
     {\mathbb{E}[\|R\|_2^2]},
\]
i.e., the fraction of total variance of that token type explained by
direction $u^{(\ell)}_j$.
Similarly, averaging $s^{(\ell)}_{\mathrm{subspace}}(i,j)$ over
high-energy tokens provides an estimate of how variance \emph{within}
the logical subspace is distributed across the $k$ directions.

\end{document}